\documentclass[10pt,twocolumn,letterpaper]{article}

\usepackage[pagenumbers]{wacv} 

\usepackage[accsupp]{axessibility}

\usepackage{listings}
\usepackage{multirow}

\definecolor{wacvblue}{rgb}{0.21,0.49,0.74}
\usepackage[pagebackref,breaklinks,colorlinks,allcolors=wacvblue]{hyperref}

\def\wacvPaperID{766} 
\def\confName{WACV}
\def\confYear{2026}

\title{SceneProp: Combining Neural Network and Markov Random Field for \\ Scene-Graph Grounding}

\author{Keita Otani\\
The University of Tokyo\\
{\tt\small otani@mi.t.u-tokyo.ac.jp}
\and
Tatsuya Harada\\
The University of Tokyo\\
RIKEN AIP\\
{\tt\small harada@mi.t.u-tokyo.ac.jp}
}

\begin{document}
\maketitle

\begin{abstract}
  Grounding complex, compositional visual queries with multiple objects and relationships is a fundamental challenge for vision-language models. 
  While standard phrase grounding methods excel at localizing single objects, they lack the structural inductive bias to parse intricate relational descriptions, often failing as queries become more descriptive.
  To address this structural deficit, we focus on scene-graph grounding, a powerful but less-explored formulation where the query is an explicit graph of objects and their relationships.
  However, existing methods for this task also struggle, paradoxically showing decreased performance as the query graph grows---failing to leverage the very information that should make grounding easier.
  We introduce SceneProp, a novel method that resolves this issue by reformulating scene-graph grounding as a Maximum a Posteriori (MAP) inference problem in a Markov Random Field (MRF).
  By performing global inference over the entire query graph, SceneProp finds the optimal assignment of image regions to nodes that jointly satisfies all constraints.
  This is achieved within an end-to-end framework via a differentiable implementation of the Belief Propagation algorithm.
  Experiments on four benchmarks show that our dedicated focus on the scene-graph grounding formulation allows SceneProp to significantly outperform prior work.
  Critically, its accuracy consistently improves with the size and complexity of the query graph, demonstrating for the first time that more relational context can, and should, lead to better grounding.
  Codes are available at \url{https://github.com/keitaotani/SceneProp}.
\end{abstract}

\section{Introduction}
\label{sec:intro}

\begin{figure}
  \centering
  \begin{subfigure}{0.99\linewidth}
    \centering
    \includegraphics[width=0.8\linewidth]{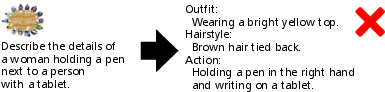}
    \caption{Image Description - GPT-4o~\cite{GPT4o} on 2025-07-10}
    \label{fig:task_gpt4o}
  \end{subfigure}
  \begin{subfigure}{0.99\linewidth}
    \centering
    \includegraphics[width=0.8\linewidth]{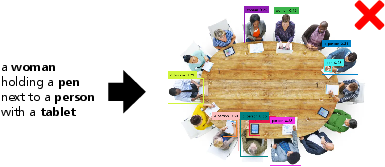}
    \caption{Phrase Grounding - Grounding DINO~\cite{GroundingDINO}}
    \label{fig:task_grounding_dino}
  \end{subfigure}
  \begin{subfigure}{0.99\linewidth}
    \includegraphics[width=0.99\linewidth]{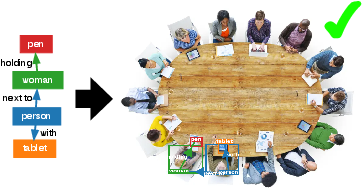}
    \caption{Scene-Graph Grounding - SceneProp (Ours)}
    \label{fig:task_sceneprop}
  \end{subfigure}
  \caption{
    Compositional grounding with SceneProp.
    Existing methods fail due to many partial matches:
    (a) GPT-4o, using description as a proxy task for its poor coordinate output, and (b) Grounding DINO.
    (c) In contrast, our SceneProp succeeds by finding an object combination that satisfies the query graph.
  }
  \label{fig:task}
\end{figure}

Visual grounding, the task of localizing objects in an image from textual descriptions, is a cornerstone of computer vision.
It is fundamental for applications ranging from robotic manipulation, such as identifying a specific item to pick up, to visual question answering, where implicitly locating objects is essential for reasoning~\cite{MDETR}.
Modern grounding models have shown remarkable capabilities, particularly in handling long-tail distributions.
Since object categories are specified in the text, the set of active categories for a given query is limited, enabling models like GLIP~\cite{GLIP} to detect even rare objects such as a "stingray".

However, the performance of these state-of-the-art methods degrades significantly when faced with compositionally complex queries that involve multiple objects and their relationships.
The more specific a query is, the easier it should be to pinpoint the target.
Yet, for a query like "a woman holding a pen next to a person with a tablet", existing models often falter (\Cref{fig:task}).
The scene contains numerous partially matching combinations—for instance, three instances of "a woman next to a person with a tablet" if the "holding a pen" clause is omitted.
This combinatorial explosion of plausible-but-incorrect candidates misleads current models.
When prompted to describe the woman, large multimodal models like GPT-4o~\cite{GPT4o} provide a conflated description of multiple individuals, while phrase grounding models like Grounding DINO 1.5~\cite{GroundingDINO1.5} often localize objects that satisfy only a subset of the specified conditions.

To address this challenge, we turn to scene-graph grounding.
This task, which takes an explicit graph of objects and relationships as a query, is a natural fit for compositional reasoning.
Despite this, the approach has remained less-explored, largely because existing methods paradoxically perform worse on more descriptive (i.e., larger) query graphs.
For instance, prior work using Graph Neural Networks (GNNs)~\cite{VLMPAG} is easily overwhelmed by this combinatorial complexity and struggles to localize objects accurately.
This critical limitation, combined with the community's historical focus on the task of Scene-Graph Generation (SGG), has left the potential of grounding underexplored.
Furthermore, prior work has largely overlooked a key advantage of the grounding formulation: its inherent robustness to long-tailed category distributions.

We propose to resolve this by reformulating scene-graph grounding as a problem of maximizing a joint probability.
For the query in \Cref{fig:task} to be correctly grounded, the joint probability of identifying the correct objects as "woman," "pen," "person," and "tablet," and their relationships as "holding," "next to," and "with," must be maximized.
This naturally leads to a Maximum a Posteriori (MAP) estimation problem over all possible assignments of image regions to graph nodes.
This probabilistic model can be represented as a Markov Random Field (MRF), allowing us to leverage sophisticated and efficient MAP inference algorithms~\cite{NIPS2007_MPLP} that can solve this seemingly complex combinatorial search.

Our method, SceneProp, first employs a neural network to estimate the unary (object) and pairwise (relationship) probabilities from visual features.
It then constructs an MRF and performs MAP inference to find the globally optimal grounding assignment.
During training, the model learns to maximize the probability of the ground-truth assignment.
Crucially, while many hybrid MRF-NN models sacrifice end-to-end differentiability, we demonstrate that Belief Propagation~\cite{reason:Pearl09a}, a classic MRF inference algorithm, can be efficiently implemented in a differentiable manner within modern define-by-run deep learning frameworks, enabling seamless end-to-end training.

We evaluated SceneProp on the VG-FO~\cite{VLMPAG}, VG150~\cite{Xu_2017_CVPR}, COCO-Stuff~\cite{COCOStuff}, and GQA~\cite{Hudson2019GQAAN} datasets, demonstrating its effectiveness.
To validate its ability to handle complex queries, we showed that on the VG-FO dataset, performance improves as the number of relationships in the query graph increases—a stark contrast to prior work.
To confirm its robustness in realistic scenarios, we showed that on the GQA dataset, with its vast number of categories, grounding performance remains stable.
This highlights a significant advantage over scene-graph generation, which suffers heavily in long-tail settings.

Our contributions are as follows:
\begin{itemize}
  \item We propose SceneProp, an end-to-end trainable method that unifies neural networks and MRFs for robust scene-graph grounding.
  \item We demonstrate that, unlike existing methods, our model's performance improves with more descriptive (i.e., larger) scene graphs.
  \item We show that our grounding performance is robust in datasets with a large number of categories, effectively handling the long-tail problem.
\end{itemize}

\section{Related Work}

\paragraph{Scene Graphs and Generation Challenges.}
Scene graphs, which represent objects and their relationships in a structured format, have been applied to a wide range of vision tasks, including visual question answering~\cite{conf/nips/HudsonM19, Qian2023SceneGR, Kim2020HypergraphAN}, image retrieval~\cite{Johnson_2015_CVPR, Wang2019CrossmodalSG}, image generation~\cite{Johnson2018ImageGF, Dhamo2020SemanticIM} and robotics~\cite{WeiYang_ICLR2019, Zhu2020HierarchicalPF}.
To power these applications, Scene-Graph Generation (SGG), the task of creating a scene graph from an image, has been the most active area of research. However, the SGG field has recently hit a bottleneck, particularly in addressing the long-tail problem~\cite{DataResampling1, Suhail2021EnergyBasedLF}.
The vast combinatorial space of object and relationship triplets in real-world data creates severe class imbalance, drastically degrading performance on less frequent categories.
In this paper, we argue that scene-graph grounding, our focus, holds a significant advantage over SGG by inherently circumventing this long-tail issue, as the categories of interest are provided in the query.

\paragraph{Visual and Scene-Graph Grounding.}
Visual grounding aims to link textual phrases to their corresponding object regions in an image.
Models like MDETR~\cite{MDETR}, GLIP~\cite{GLIP} and Grounding DINO~\cite{GroundingDINO,GroundingDINO1.5} excel at this, even for open-vocabulary or long-tail objects, by leveraging textual cues.
However, they often fail to parse compositionally complex queries with multiple interrelated objects, as they lack an explicit inductive bias for graph structures.

Scene-graph grounding directly addresses this by using a graph as the query.
Early work used hand-crafted features~\cite{Johnson_2015_CVPR}, but the state-of-the-art is represented by VL-MPAG~\cite{VLMPAG}, an end-to-end neural model.
While effective, VL-MPAG's performance paradoxically degrades as the query graph becomes more descriptive (i.e., larger).
This is due to its 2-layer GNN, which cannot capture long-range dependencies and suffers from over-smoothing if layers are added.
Our method overcomes this by performing global inference, ensuring that more relationships lead to better accuracy.

\begin{figure}
  \centering
  \includegraphics[width=0.99\linewidth]{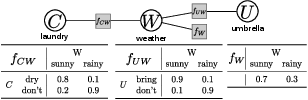}
  \caption{
    A simple example of an MRF.
  }
  \label{fig:mrf_weather_example}
\end{figure}

\paragraph{MRFs with Neural Networks.}
A Markov Random Field (MRF) models a joint probability distribution over a set of variables using a graph.
For instance, in \Cref{fig:mrf_weather_example}, the probability of observing laundry ($C$), weather ($W$), and an umbrella ($U$) can be expressed as a product of potential functions on the graph's cliques:
\begin{equation}
  P(c,w,u) = \frac{1}{Z} f_{CW}(c,w)f_{UW}(u,w)f_W(w)
\end{equation}
where Z is a normalization constant.
This structure allows for efficient probabilistic inference, such as MAP estimation, using well-established algorithms~\cite{reason:Pearl09a, NIPS2007_MPLP}.

Research combining neural networks with MRFs has been conducted in semantic segmentation~\cite{Chen2016DeepLabSI, brudfors2021mrf}, community detection~\cite{jin2019graph, he2021community}, and feature point matching~\cite{guan2024neural, liang2023neural}.
A key strength of MRFs is that non-adjacent variables can be correlated through intermediates, but this same property complicates differentiation for end-to-end learning.
Consequently, some methods either forgo end-to-end training and use MRFs only for inference~\cite{Chen2016DeepLabSI, brudfors2021mrf, liang2023neural}, or restrict correlations to only adjacent variables~\cite{jin2019graph}.
Other works use GNNs inspired by MRF algorithms~\cite{he2021community, guan2024neural}, but these approaches do not retain the probabilistic nature of MRFs and inherit GNN-specific limitations like over-smoothing.
In contrast, we find that the classic Belief Propagation~\cite{reason:Pearl09a} algorithm can be efficiently implemented in define-by-run deep learning libraries in a manner that is automatically differentiable.
This enables true end-to-end learning while preserving the full probabilistic inference capabilities of the MRF.

\begin{figure*}
  \centering
  \includegraphics[width=0.99\linewidth]{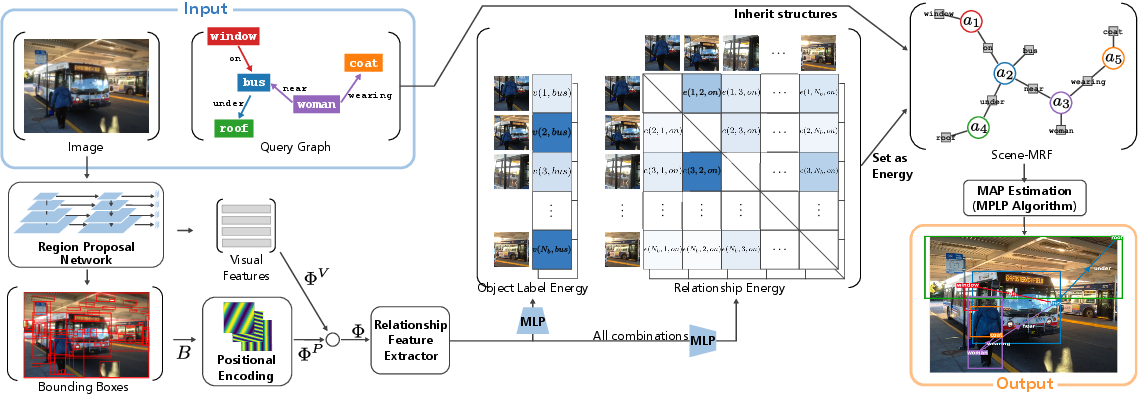}
  \caption{
    Overview of the SceneProp's pipeline.
    First, image features and bounding boxes are extracted (\Cref{sec:region_proposal}).
    The bounding boxes are encoded using our proposed positional encoding (\Cref{sec:positional_encoding}).
    Using these encoded positions and image features, the relationship features are extracted (\Cref{sec:relationship_feature}).
    Then, the energy of the object label and relationship are calculated using the relationship feature.
    Using this energy and the given query graph, we construct and infer the Scene-MRF (\Cref{sec:inference_with_scene_mrf}).
  }
  \label{fig:model}
\end{figure*}

\section{Method}

SceneProp takes advantage of both neural networks and MRFs by combining them in an end-to-end manner. 
An overview of this process is presented in \Cref{fig:model}.

\subsection{Mathematical Notation and Variables}
A query graph, which is a scene graph without links to an image, is a graph structure in which nodes represent objects and edges represent relationships between objects.
Nodes are assigned object names (e.g., apple), and edges are assigned relationships (e.g., on).
Formally, we assume that there are $N$ objects in the query graph, with IDs assigned from 1 to $N$.
The object label is represented by $O=\{(i_1, o_1), (i_2, o_2), \ldots, (i_m, o_m)\}$.
Here, $i_1,\ldots, i_m$ indicates the object's ID, and $o_1, \ldots, o_m$ indicates the object label's ID.
The relationship label is represented by $R=\{(j_1, k_1, r_1), (j_2, k_2, r_2), \ldots, (j_n, k_n, r_n)\}$.
Here, $j_1,\ldots, j_n$, $k_1,\ldots, k_n$ indicates the object's ID, and $r_1,\ldots, r_n$ indicate the relationship label's ID.
Then, the query graph is represented as $G=(O, R)$.

For the image $I$, the bounding boxes of the objects are listed as $B=\{b_1, b_2 , \ldots, b_{N_b}\}$.
We assume that an object with ID $i$ in the query graph corresponds to a bounding box with ID $a_i$.
Subsequently, the set of objects assigned to the boxes is $A=\{a_1, a_2, \ldots, a_{N}\}$.

The goal of scene-graph grounding is to estimate the bounding boxes $B$ and assignment $A$ of the nodes for a given image $I$ and query graph $G$.

\subsection{Scene-MRF Formulation}
\label{sec:mrf_formulation}
We explain the MRF formulation using a simple query graph \verb'cup-[on]-table'.
Let $O=\{(1, {\rm cup}), (2, {\rm table})\}$, $R=\{(1, 2, {\rm on})\}$, and $A=\{a_1, a_2\}$, where the goal is to select $a_1, a_2$ from the set of object candidate boxes $B$.
First, selecting \verb'cup' from the set of object candidate boxes $B$ is equivalent to sampling from the probability distribution $P(a_1|{\rm cup})$.
The same applies to \verb'table'.
Next, selecting a pair of object candidate boxes with the relationship \verb'[on]' is equivalent to sampling from the probability distribution $P(a_1, a_2|{\rm on})$.
By integrating these, the distribution for selecting $a_1, a_2$ from the set of object candidate boxes $B$ is defined as follows:
\begin{equation}
  P(a_1, a_2|G, I) = \frac{1}{Z} P(a_1|{\rm cup})P(a_2|{\rm table})P(a_1, a_2|{\rm on})
\end{equation}
where $Z$ is the normalization term.

For a general query graph, we similarly define the following probability distribution:
\begin{equation}
  P(A|G, I) = \frac{1}{Z} \prod_{i, o \in O} P(a_i|o) \prod_{j, k, r \in R} P(a_j, a_k|r)
\end{equation}
Taking the logarithm and defining $-\log P(a_i|o)$ as $v(a_i, o)$ and $-\log P(a_j, a_k|r)$ as $e(a_j, a_k, r)$, we can express it as follows:
\begin{gather}
  \label{eq:scenemrf_nll}
  -\log P(A|G, I) = \sum_{i, o \in O} v(a_i, o) + \sum_{j, k, r \in R} e(a_j, a_k, r) + Z \\
  Z = \log \sum_A \exp \left(-\sum_{i, o \in O} v(a_i, o) - \sum_{j, k, r \in R} e(a_j, a_k, r)\right)
\end{gather}
We call the MRF representing this probability distribution the Scene-MRF, which has the same structure as the query graph, as shown in the upper right of \Cref{fig:model}.
The terms $v(a_i, o)$ and $e(a_j, a_k, r)$ are called energies and are estimated using neural networks, as explained in later sections.

\subsection{Region Proposal and Visual Feature Extraction}
\label{sec:region_proposal}
We use ATSS~\cite{Zhang2019BridgingTG} to estimate the bounding boxes $B=\{b_1, b_2, \ldots, b_{N_b}\}$ of the object candidate regions from image $I$.
We use a network that connects the Swin transformer~\cite{Liu2021SwinTH} and the Feature Pyramid Network~\cite{Lin2016FeaturePN} to extract visual features from the image.
We also extract the visual features of the bounding boxes, resulting in $\Phi^V=\{\phi^V_1, \phi^V_2, \ldots, \phi^V_{N_b}\}$.

\subsection{Positional Encoding}
\label{sec:positional_encoding}
The positional relationship between objects is important for scene-graph grounding.
Sinusoidal encoding~\cite{Vaswani2017AttentionIA} is commonly used for positional information but is limited to one-dimensional points and cannot encode spatial ranges like bounding boxes.
Therefore, we developed a new positional encoding method to represent two-dimensional positional ranges.
A 2D range with a center coordinate of $(x,y)$ and width and height of $(\sigma_x,\sigma_y)$ is encoded as follows:
\begin{gather}
  pos^\sigma_i = \exp(-\sigma_x(1-\cos \theta^x_i)-\sigma_y(1-\cos\theta^y_i)) \\
  pos^{cos}_i = pos^\sigma_i\cos(\theta^x_i x+\theta^y_i y) \\
  pos^{sin}_i = pos^\sigma_i\sin(\theta^x_i x+\theta^y_i y) \\
  pos = {\rm concat}(pos^{cos}_1, \ldots, pos^{cos}_n, pos^{sin}_1, \ldots, pos^{sin}_n)
\end{gather}
$pos$ denotes the encoding results.
Here, $\theta^x_i, \theta^y_i$ are integers and $-\theta_{max}\leq\theta^x_i\leq\theta_{max}, 0\leq\theta^y_i\leq\theta_{max}$.
Using all $(\theta_i^x,\theta_i^y)$ combinations is desirable but leads to a prohibitively high-dimensional $pos$ vector.
We therefore keep all pairs with $(\theta_i^x)^2+(\theta_i^y)^2 \le (\theta_{\text{low}})^2$ and preselect a random subset of the remainder that is held fixed during training and inference.

Our positional encoding method uses Fourier transforms to effectively encode positional relationships.
By applying an inverse Fourier transform, we derive a function centered at $(x, y)$ with a width and height of $(\sigma_x, \sigma_y)$, (see the supplementary material).
This method captures spatial ranges and allows parallel movement and range expansion through simple multiplication operations.
The inner product of the encoded results reflects the degree of overlap between ranges, as the Fourier transform preserves inner products.

We apply positional encoding to encode the bounding boxes $B=\{b_1, b_2, \ldots, b_{N_b}\}$, resulting in $\Phi^P=\{\phi^P_1, \phi^P_2, \ldots, \phi^P_{N_b}\}$.
In this process, we use the box width and height for $(\sigma_x, \sigma_y)$.
We use $\theta_{max}=48, \theta_{low}=6$, and select 128 combinations of $\theta^x_i, \theta^y_i$.

\subsection{Extraction of Relationship Features}
\label{sec:relationship_feature}
To recognize the relationships between objects, it is important to propagate object features mutually.
Various methods have been proposed in the field of visual relationship detection that employ RNN, GNN, and transformers to mutually propagate object features and extract inter-object relationships~\cite{Xu_2017_CVPR, GraphRCNN, Tang_2019_CVPR_VCTREE_mRaN, TransformerSGG1, TransformerSGG2}.
Following these approaches, we use a 2-layer transformer encoder ~\cite{Vaswani2017AttentionIA} to mutually propagate the visual features of the object $\Phi^V$ and positional encoding $\Phi^P$.
This operation yields relationship features $\Phi$ that consider visual features, positional features, and inter-object relationships.

\begin{eqnarray}
  \phi'_i &=& {\rm concat}(\phi^V_i, \phi^P_i) \quad \forall i \in \{1, \ldots , N_b\} \\
  \Phi &=& \{\phi_1, \phi_2, \ldots, \phi_{N_b}\} \nonumber \\
  &=& {\rm Transformer}(\{\phi'_1, \phi'_2, \ldots, \phi'_{N_b}\})
\end{eqnarray}

\subsection{Inference with Scene-MRF}
\label{sec:inference_with_scene_mrf}
Calculate the Scene-MRF energies $v(a_i, o)$ and $e(a_j, a_k, r)$ defined in \Cref{sec:mrf_formulation} to complete the Scene-MRF.
For reference, $v(a_i, o)$ is the energy when object $i$ is assigned label $o$, and $e(a_j, a_k, r)$ is the energy when objects $j$ and $k$ are assigned relationship $r$.
$v(a_i, o)$ is the $o$th element of the vector obtained by passing the feature $\phi_{a_i}$ of object $a_i$ through a 2-layer multilayer perceptron (MLP).
Similarly, $e(a_j, a_k, r)$ is the $r$-th element of the result of concatenating the features of objects $a_j$ and $a_k$ and processing them using a 3-layer MLP.
By substituting this into \Cref{eq:scenemrf_nll}, the probability distribution $P(A|G, I)$ of the assignment $A$ is obtained.

Two inference methods are used in this study.
The first method finds the optimal assignment combination for all objects, which is ideal for listing the grounding results.
The second method determines the optimal assignment for a specific object that is suitable for localizing it.

In the first method, the grounding result $A^*$ is the maximum a posteriori (MAP) estimation of the Scene-MRF as shown in the following equation:
\begin{equation}
  A^* = \mathop{\rm arg~max}\limits_A \log P (A|G, I)
\end{equation}
Finding such $A^*$ is generally NP-hard, so we use the MPLP algorithm~\cite{NIPS2007_MPLP} for efficient approximation.
The MPLP algorithm can provide an exact solution when a graph has a tree structure.
In the MAP estimation results, different objects may be assigned to the same bounding box.
To address this, we add the constraint that different objects should be assigned to different bounding boxes and fine-tune the grounding results using the Markov Chain Monte Carlo method (MCMC) with the MPLP algorithm result as the initial value.

In the second method, the grounding result for a specific object $a_i$ is obtained as the marginal probability distribution $P(a_i|G, I)$:
\begin{equation}
  P(a_i|G, I) = \sum_{A \backslash a_i} P(A|G, I)
\end{equation}
If a query graph has a tree structure, the exact solution can be calculated efficiently using belief propagation~\cite{reason:Pearl09a}.
Otherwise, it can be approximated using a generalized belief propagation algorithm~\cite{reason:Pearl09a}.

\subsection{Training}
\label{sec:training}
The grounding loss $L_g$ is computed using the ground-truth assignment $A^{gt}$ as shown in the following equation:
\begin{equation}
  \begin{aligned}
    L_g &= -\log P (A^{gt}|G, I) \\
    &= \sum_{i, o \in O} v(a^{gt}_i, o) + \sum_{j, k, r \in R} e(a^{gt}_j, a^{gt}_k, r) + Z
    \label{eq:loss}
  \end{aligned}
\end{equation}
For tree-structured MRFs, belief propagation~\cite{reason:Pearl09a} efficiently computes $Z$.
This algorithm's operations are differentiable, enabling end-to-end training and automatic differentiation in deep-learning libraries.
Because belief propagation cannot handle non-tree graphs, we randomly extracted tree-structured subgraphs and trained on them.
Random weights were assigned to the edges, and we extracted the subgraph by finding the minimum spanning tree using Kruskal's algorithm~\cite{Kruskal1956}.
Belief propagation plays a crucial role in our method, and detailed information is provided in the supplementary material.

The loss for the object candidate region extraction, $L_b$, is computed by following the method used in ATSS~\cite{Zhang2019BridgingTG}.
The total loss is then calculated as $L = L_g + L_b$.

SceneProp can learn grounding that considers the entire graph.
For query graphs with some partial matches, the object locations are expected to be specified in a chain starting from the most easily identifiable object.
SceneProp learns chain-like grounding using belief propagation, because belief propagation is a message-passing algorithm that allows information to propagate from one end of a graph to another.
This contrasts with VL-MPAG~\cite{VLMPAG}, which only propagates information 2-hop.

\begin{table*}
  \centering
  \begin{tabular}{ll|rr|rr|rr|rr|rr|r}
    \toprule
    Type & Model & \multicolumn{2}{|c|}{VG-FO} & \multicolumn{2}{|c|}{VG150} & \multicolumn{2}{|c|}{COCO-Stuff} & \multicolumn{2}{|c|}{GQA} & \#params \\
    & & R@1 & R@5 & R@1 & R@5 & R@1 & R@5 & R@1 & R@5 & \\
    \midrule
    LVLMs & Florence-2-L~\cite{florence2}         & 19.6 & - & 19.0 & - & 18.5 & - & 10.2 & - & 770M \\
    \midrule
    Phrase & MDETR~\cite{MDETR}                   & 25.4 & 44.8 & 27.0 & 45.3 & 30.1 & 47.9 & - & - & 169M \\
    Grounding & GLIP-T~\cite{GLIP}                & 31.1 & 51.3 & 30.4 & 50.3 & 45.9 & 80.4 & 23.6 & 49.8 & 232M \\
    & GroundingDINO~\cite{GroundingDINO}          & 32.0 & 48.5 & 31.6 & 47.9 & 47.3 & 80.4 & 25.4 & 47.0 & 172M \\
    \midrule
    Scene-Graph & VL-MPAG~\cite{VLMPAG} (w/o rels) & 29.9 & 53.5 & - & - & 33.9 & 57.2 & - & - & 69M \\
    Grounding & VL-MPAG~\cite{VLMPAG}             & 36.0 & 63.3 & 33.6 & 59.9 & 36.3 & 58.4 & 5.2 & 18.7 & 69M \\
    & SceneProp (Ours, w/o rels)                   & \underline{40.0} & \underline{64.6} & \underline{37.4} & \underline{64.1} & 48.1 & \underline{82.8} & 33.4 & 54.5 & \textbf{34M} \\
    & SceneProp (Ours, w/o an obj)                   & 32.0 & 53.0 & 30.4 & 51.6 & \underline{53.1} & 81.9 & \underline{36.5} & \underline{55.1} & \textbf{34M} \\
    & SceneProp (Ours)                            & \textbf{46.6} & \textbf{67.1} & \textbf{43.7} & \textbf{68.4} & \textbf{68.2} & \textbf{88.9} & \textbf{53.6} & \textbf{68.2} & \textbf{34M} \\
    \bottomrule
  \end{tabular}
  \caption{
    Comparison of Recall@1 and Recall@5 for VG-FO, VG-150, COCO-Stuff, and GQA.
    Bold values indicate the best and underlined values indicate the second-best performance.
    (w/o rels) denotes results without considering relationships.
    For VL-MPAG~\cite{VLMPAG}, VG150 and GQA results are unavailable as the (w/o rels) implementation is not provided.
    For phrase grounding models, the query graph is flattened like "apple on table. cup on table." and fine-tuned as in \cite{VLMPAG}.
    For Florence-2-L~\cite{florence2}, "Locate the phrases in the caption:" with flattened query graphs is used as the prompt.
  }
  \label{tab:main_result}
\end{table*}

\begin{table}
  \centering
  \begin{tabular}{l|rrrr}
    \toprule
    Dataset & R@1 & R@5 & mR@1 & mR@5 \\
    \midrule
    VG-FO      & 31.3 & 46.6 & 27.9 & 43.1 \\
    VG150      & 29.4 & 42.8 & 29.6 & 43.3 \\
    COCO-Stuff & 46.8 & 70.0 & 48.5 & 71.0 \\
    GQA        & 32.5 & 46.7 & 39.0 & 53.2 \\
    \bottomrule
  \end{tabular}
  \caption{
    Recall@1, Recall@5, mean Recall@1, and mean Recall@5 for object pairs with relationships.
  }
  \label{tab:pair_eval}
\end{table}

\section{Experiments}

\subsection{Dataset and Evaluation Protocol}
We used VG-FO~\cite{VLMPAG}, VG150~\cite{Xu_2017_CVPR}, COCO-Stuff~\cite{COCOStuff}, and GQA~\cite{Hudson2019GQAAN} for evaluation.
VG-FO and VG150 are datasets derived from the Visual Genome with reduced numbers of objects and relationship categories.
VG-FO was proposed in VL-MPAG~\cite{VLMPAG} for scene-graph grounding and includes 150 object and 40 relationship categories, with 93K training and 40K test images.
VG150 is the most widely used subset in the scene-graph community, containing 150 object and 50 relationship categories, with 108K training and 10K test images.
The COCO-Stuff dataset originally used for panoptic segmentation was converted into scene graphs using the rule-based method in \cite{Johnson2018ImageGF}.
We also used the GQA dataset, which is well known in the field of scene-graph generation, for performance evaluation.
For COCO-Stuff and GQA, the counts of the training images, test images, object categories, and relationship categories were (48K, 2K, 183, 6) and (74K, 11K, 1981, 331), respectively.
The GQA dataset, with its large number of object and relationship categories, closely resembles the actual scenes.

We considered the intersection over union (IoU) of the ground-truth bounding box and the estimated result to be correct if it exceeded 0.5.
The evaluation metrics were recall@1 (equivalent to accuracy) and recall@5.

\subsection{Implementation Details}
We used a Swin transformer+FPN~\cite{Liu2021SwinTH, Lin2016FeaturePN} pre-trained on the Object365~\cite{Objects365} dataset as the backbone network.
There were 512 object candidates obtained from the object detector.
Data augmentation included random resizing, Gaussian noise addition, and color jittering.
To improve relationship learning, we randomly replaced object categories with the "object" category with a probability of 0.2 during training.
We used the Adam optimizer with a learning rate of 1e-4 and a batch size of 20.
Training was conducted on two A100 40 GB GPUs for 36 h.
For a quantitative evaluation, we calculated recall by determining the marginal distribution of the object assignment using belief propagation.
Unseen categories in the VG-PO dataset and unlabeled objects in \Cref{sec:analysis_of_properties} were treated as the "object" category.
The inference time for 6,340 images from the VG-FO dataset was 32 min using a single A100 40 GB GPU.
For comparison, VL-MPAG~\cite{VLMPAG} takes 36 h for training and 43 min for inference using a single A100 40 GB GPU.

\subsection{Main Quantitative Results}
\label{sec:main_results}

We evaluate SceneProp against a range of relevant grounding models.
Our baselines include Florence-2-L~\cite{florence2}, an LVLM capable of precise object localization, leading phrase grounding models, and the previous state-of-the-art in scene-graph grounding, VL-MPAG~\cite{VLMPAG}.
We chose Florence-2-L as many other prominent LVLMs, such as GPT-4o, lack reliable coordinate outputs for this task.
Newer phrase grounding models, including GroundingDINO1.5 \& 1.6~\cite{GroundingDINO1.5}, DetCLIPv3~\cite{detclipv3}, and T-Rex2~\cite{TRex2}, were not included in our comparison as their official code or pre-trained models were unavailable at the time of our experiments.
\Cref{tab:main_result} shows that SceneProp significantly outperforms all baselines across four diverse datasets, despite having the smallest model size.
Notably, on the complex GQA dataset, which features large query graphs and a vast number of categories, SceneProp achieves a remarkable +48.4\% R@1 improvement over VL-MPAG.

\begin{figure}
  \centering
  \includegraphics[width=0.99\linewidth]{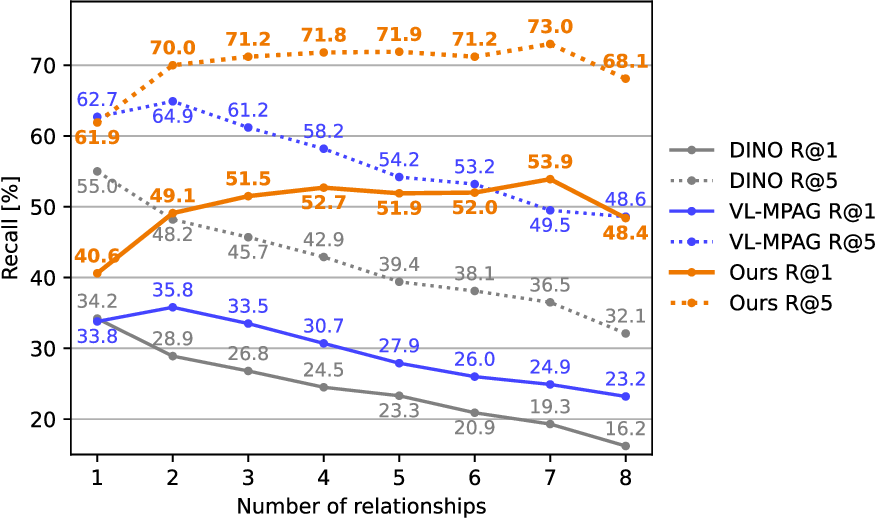}
  \caption{
    Impact of query graph size (number of relationships) on grounding performance on the VG-FO dataset. 
    SceneProp's performance improves with more descriptive queries, while both VL-MPAG and Grounding DINO (DINO) show declining recall.
  }
  \label{fig:n_relation}
\end{figure}

This superior performance stems directly from our core proposal: formulating grounding as a global inference problem on an MRF.
Unlike methods based on local message passing, such as VL-MPAG's 2-layer GNN, SceneProp's performance scales with the descriptive power of the query graph.
\Cref{fig:n_relation} validates this key hypothesis.
As the number of relationships in the query increases, VL-MPAG's performance degrades due to its limited receptive field.
In stark contrast, SceneProp's accuracy improves, as more relational constraints help the MRF inference to better disambiguate the globally optimal assignment.
This effect is most pronounced on GQA, where the average query graph has 9.5 relationships, causing baselines to falter while SceneProp excels.

For comparison, we adapted phrase grounding models for the scene-graph grounding task.
Following the methodology of VL-MPAG~\cite{VLMPAG}, we flattened the query graph into a sentence (e.g., "apple on table. cup on table.") and fine-tuned the models.
We also tried a structure-preserving format with object indices (e.g., "apple 1 on table 1. cup 2 on table 1.") but found it degraded performance.
While these models achieve reasonable scores, they share the same fundamental weakness as VL-MPAG.
As \Cref{fig:n_relation} shows, their performance also declines with larger query graphs, confirming that they struggle with compositional queries due to a lack of an explicit inductive bias for graph structures.

\begin{figure*}
  \centering
  \includegraphics[width=0.99\linewidth]{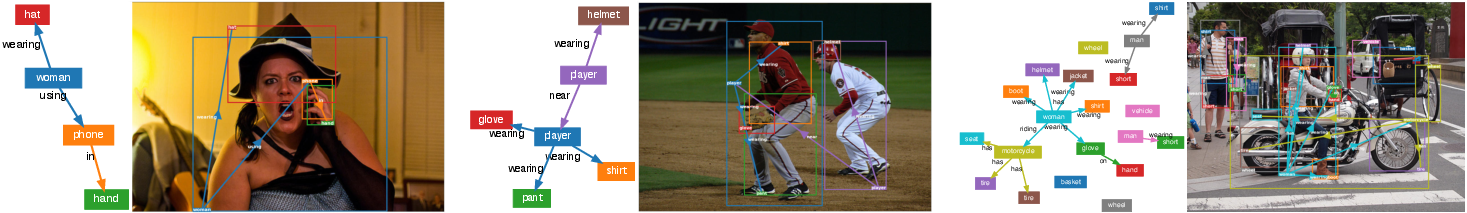}
  \caption{
    Examples of SceneProp's inference results.
    The query graph for each image is shown with colored nodes that correspond to the bounding boxes in the output grounding.
    Our model successfully localizes all objects and their relationships even in complex scenes.
    Note that the "Output" image in the pipeline diagram (\Cref{fig:model}) is also an actual inference result from our method.
    More examples are available in the supplementary material.
  }
  \label{fig:example}
\end{figure*}

\begin{figure}
  \centering
  \includegraphics[width=0.99\linewidth]{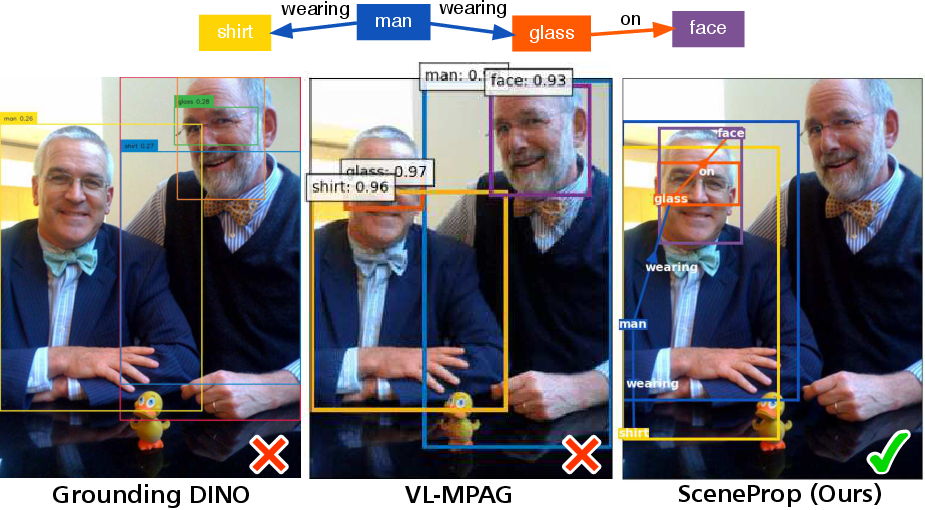}
  \caption{
    Qualitative comparison.
    Unlike baselines such as Grounding DINO and VL-MPAG that are misled by partial matches, SceneProp correctly grounds all objects by satisfying the full set of relational constraints.
  }
  \label{fig:vlmpag_fail_example}
\end{figure}

Qualitative examples of SceneProp's ability to resolve complex queries are shown in \Cref{fig:example} and \Cref{fig:vlmpag_fail_example}.

\subsection{Analyzing SceneProp's Properties}
\label{sec:analysis_of_properties}
\begin{figure}
  \centering
  \begin{subfigure}{0.99\linewidth}
    \includegraphics[width=0.72\linewidth]{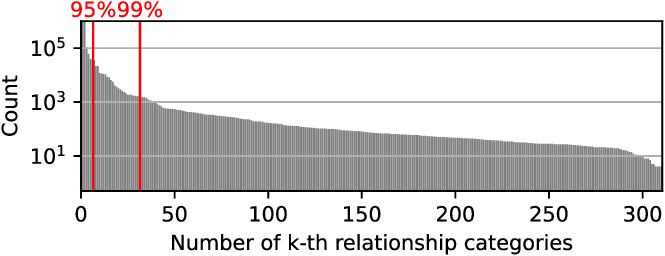}
  \end{subfigure}
  \begin{subfigure}{0.99\linewidth}
    \hspace{0.7mm}
    \includegraphics[width=0.88\linewidth]{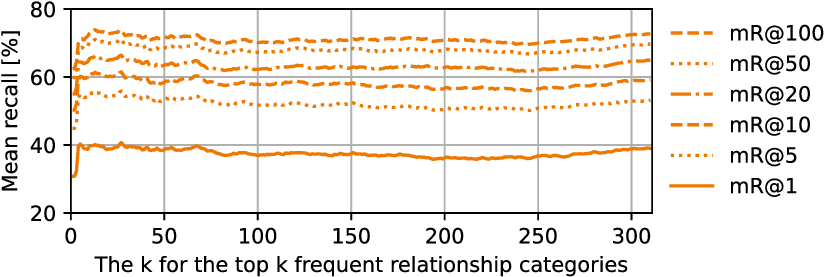}
  \end{subfigure}
  \caption{
    Mean recall over top-k frequent relationships on GQA.
    SceneProp's performance remains stable even when including rare relationship categories.
  }
  \label{fig:gqa_recall_topk_freq_results}
\end{figure}

\paragraph{Robustness on Long-Tail Relationships.}
A central claim of our work is that grounding mitigates the long-tail issue in scene-graph generation.
We therefore introduce a stricter pair-centric evaluation where a relationship is counted correct only if \emph{both objects} are correctly grounded, and we report Recall (R@$k$) and mean Recall (mR@$k$) that averages recall per relationship category.
Across datasets, \Cref{tab:pair_eval} shows a small R@$k$-mR@$k$ gap.
On GQA, we compute mR over the top-$k$ most frequent relationship categories (\Cref{fig:gqa_recall_topk_freq_results}); a near-flat curve indicates stability as rarer categories are included, evidencing robustness to long-tail categories.
By contrast, scene-graph generation methods typically struggle under GQA's pronounced long-tail, with mR degrading as rarer categories enter; see the supplementary material for detailed comparisons.

\begin{table}
  \centering
  \begin{tabular}{rr|rr}
    \toprule
    \multicolumn{2}{c|}{w/o loops} & \multicolumn{2}{c}{w/ loops} \\
    R@1 & R@5 & R@1 & R@5 \\
    \midrule
    45.7 & 66.3 & 58.1 & 77.6 \\
    \bottomrule
  \end{tabular}
  \caption{
    VG-FO performance on query graphs w/o vs.\ w/ loops.
  }
  \label{tab:without_objlabel}
\end{table}

\paragraph{Generalization from Trees to Loops.}
We evaluate the effect of loops in the query graph on VG-FO (\Cref{tab:without_objlabel}).
Despite training exclusively on tree-structured subgraphs for exact belief propagation, SceneProp performs better on loopy graphs.
This indicates that SceneProp generalizes to cycles and benefits from their richer relational constraints (typically more relationships per object).

\paragraph{Role of Relations and Labels in Grounding.}
To assess the role of relational reasoning, we first remove all relational edges (\Cref{tab:main_result}, w/o rels) and observe a substantial performance drop, especially on GQA where query graphs contain many relations, highlighting the critical importance of relational information.
We then mask one object's category label at a time while keeping all relations and the other labels fixed, and evaluate grounding only for the masked target (\Cref{tab:main_result}, w/o an obj).
Additionally, the supplementary evaluates grounding with all object labels entirely removed—using only relations.

\subsection{Ablation Study}
\label{sec:ablation}
We conducted ablation studies on VG-FO to validate our design choices (\Cref{tab:posenc_result}).
\textbf{(1) Architecture:} Using the same ResNet50-FPN backbone, our method still outperforms VL-MPAG, highlighting the strength of our MRF-based approach.
Removing the relationship feature extractor or our positional encoding also hurts performance, confirming their contributions.
\textbf{(2) MRF Formulation:} Replacing the entire MRF with a simple MLP classifier (``w/o MRF'') drops R@1 by 6.6\%, showing that global inference is critical.
\textbf{(3) Training:} We compare end-to-end training using belief propagation (BP) with pseudo-likelihood (PL)~\cite{PseudoLikelihood} and generalized BP (GBP; a variational surrogate)~\cite{reason:Pearl09a}.
GBP handles loopy graphs, but its variational surrogate hurts performance because the optimum of its biased objective diverges from the true solution; PL is asymptotically consistent with BP~\cite{PseudoLikelihood}, yet with limited data it fails to recover the true energy and yields lower accuracy.
In contrast, BP optimizes the exact loss on tree-structured graphs and outperforms both, validating our choice.

\begin{table}
  \centering
  \begin{tabular}{ll|rr}
    \toprule
    \multicolumn{2}{l|}{Method} & R@1 & R@5 \\
    \midrule
    Model & w/ same backbone as ~\cite{VLMPAG} & 41.5 & 64.6 \\
    & w/o Relationship extractor & 44.2 & 65.1 \\
    & w/o Positional Encoding & 43.4 & 64.9 \\
    & w/ Transformer PosEnc & 45.7 & 66.9 \\
    & w/o MRF & 40.0 & 64.6 \\
    \midrule
    Loss & w/ GBP & 43.0 & 64.6 \\
    & w/ Pseudo-likelihood V & 45.3 & 67.1 \\
    & w/ Pseudo-likelihood E & 45.3 & \textbf{67.2} \\
    \midrule
    \multicolumn{2}{l|}{SceneProp (Ours)} & \textbf{46.6} & 67.1 \\
    \bottomrule
  \end{tabular}
  \caption{
    Ablation study on the VG-FO dataset. Our proposed components and training strategy are crucial for optimal performance.
  }
  \label{tab:posenc_result}
\end{table}

\subsection{Limitations and Future Work}
\label{sec:limitations}
While SceneProp is effective within its current design, we identify two primary areas for future extension.
First, the model currently operates on a closed-set vocabulary, limiting its use to pre-defined categories.
Extending it to an open-vocabulary setting is a crucial next step to broaden its real-world applicability.
Second, our method relies on a structured graph as input.
A promising future direction is to develop a hybrid system that leverages modern LVLMs to parse free-form text into a graph, thus combining the flexibility of natural language with SceneProp's strong relational reasoning.

\section{Conclusion}
\label{sec:conclusion}
We introduced SceneProp, a novel method that reformulates scene-graph grounding as a MAP inference problem in a Markov Random Field.
By combining the feature extraction power of neural networks with the global reasoning capability of MRFs in an end-to-end trainable framework, SceneProp overcomes a key limitation of prior work: performance degradation on complex, descriptive queries.
Our experiments show that SceneProp's accuracy scales with query complexity and is robust to the long-tail distributions common in real-world data.
By demonstrating a principled way to handle compositional visual queries, we hope to spur further research into scalable and robust scene-graph grounding.

\section*{Acknowledgements}
This work was partially supported by JST Moonshot R\&D Grant Number JPMJPS2011, CREST Grant Number JPMJCR2015 and Basic Research Grant (Super AI) of Institute for AI and Beyond of the University of Tokyo.

{
  \small
  \bibliographystyle{ieeenat_fullname}
  \bibliography{egbib}

@String(CVPR= {IEEE Conf. Comput. Vis. Pattern Recog.})

@String(ICCV= {Int. Conf. Comput. Vis.})

@String(ECCV= {Eur. Conf. Comput. Vis.})

@String(ICLR = {Int. Conf. Learn. Represent.})

@String(AAAI = {AAAI})

@String(CVPR  = {CVPR})

@String(ICCV  = {ICCV})

@String(ECCV  = {ECCV})

@String(ICLR  = {ICLR})

@book{reason:Pearl09a,
  author = {Pearl, Judea},
  edition = {2nd},
  publisher = {Cambridge University Press},
  title = {Causality: Models, Reasoning and Inference},
  year = 2009
}

@inproceedings{NIPS2007_MPLP,
 author = {Globerson, Amir and Jaakkola, Tommi},
 booktitle = {Advances in Neural Information Processing Systems},
 editor = {J. Platt and D. Koller and Y. Singer and S. Roweis},
 pages = {},
 publisher = {Curran Associates, Inc.},
 title = {Fixing Max-Product: Convergent Message Passing Algorithms for MAP LP-Relaxations},
 url = {https://proceedings.neurips.cc/paper_files/paper/2007/file/8d6dc35e506fc23349dd10ee68dabb64-Paper.pdf},
 volume = {20},
 year = {2007}
}

@article{PseudoLikelihood,
 ISSN = {00390526, 14679884},
 URL = {http://www.jstor.org/stable/2987782},
 author = {Julian Besag},
 journal = {Journal of the Royal Statistical Society. Series D (The Statistician)},
 number = {3},
 pages = {179--195},
 publisher = {[Royal Statistical Society, Wiley]},
 title = {Statistical Analysis of Non-Lattice Data},
 urldate = {2024-09-08},
 volume = {24},
 year = {1975}
}

@INPROCEEDINGS{VLMPAG,
  author={Tripathi, Aditay and Mishra, Anand and Chakraborty, Anirban},
  booktitle={2023 IEEE/CVF Winter Conference on Applications of Computer Vision (WACV)}, 
  title={Grounding Scene Graphs on Natural Images via Visio-Lingual Message Passing}, 
  year={2023},
  volume={},
  number={},
  pages={4380-4389},
  doi={10.1109/WACV56688.2023.00437}}

@InProceedings{Johnson_2015_CVPR,
author = {Johnson, Justin and Krishna, Ranjay and Stark, Michael and Li, Li-Jia and Shamma, David and Bernstein, Michael and Fei-Fei, Li},
title = {Image Retrieval Using Scene Graphs},
booktitle = {Proceedings of the IEEE Conference on Computer Vision and Pattern Recognition (CVPR)},
month = {June},
year = {2015}
}

@article{Chen2016DeepLabSI,
  title={DeepLab: Semantic Image Segmentation with Deep Convolutional Nets, Atrous Convolution, and Fully Connected CRFs},
  author={Liang-Chieh Chen and George Papandreou and Iasonas Kokkinos and Kevin P. Murphy and Alan Loddon Yuille},
  journal={IEEE Transactions on Pattern Analysis and Machine Intelligence},
  year={2016},
  volume={40},
  pages={834-848},
  url={https://api.semanticscholar.org/CorpusID:3429309}
}

@inproceedings{brudfors2021mrf,
  title={An MRF-UNet product of experts for image segmentation},
  author={Brudfors, Mikael and Balbastre, Ya{\"e}l and Ashburner, John and Rees, Geraint and Nachev, Parashkev and Ourselin, S{\'e}bastien and Cardoso, M Jorge},
  booktitle={Medical Imaging with Deep Learning},
  pages={48--59},
  year={2021},
  organization={PMLR}
}

@inproceedings{jin2019graph,
  title={Graph convolutional networks meet markov random fields: Semi-supervised community detection in attribute networks},
  author={Jin, Di and Liu, Ziyang and Li, Weihao and He, Dongxiao and Zhang, Weixiong},
  booktitle={Proceedings of the AAAI conference on artificial intelligence},
  volume={33},
  number={01},
  pages={152--159},
  year={2019}
}

@inproceedings{he2021community,
  title={Community-centric graph convolutional network for unsupervised community detection},
  author={He, Dongxiao and Song, Yue and Jin, Di and Feng, Zhiyong and Zhang, Binbin and Yu, Zhizhi and Zhang, Weixiong},
  booktitle={Proceedings of the twenty-ninth international conference on international joint conferences on artificial intelligence},
  pages={3515--3521},
  year={2021}
}

@inproceedings{guan2024neural,
  title={Neural markov random field for stereo matching},
  author={Guan, Tongfan and Wang, Chen and Liu, Yun-Hui},
  booktitle={Proceedings of the IEEE/CVF Conference on Computer Vision and Pattern Recognition},
  pages={5459--5469},
  year={2024}
}

@article{liang2023neural,
  title={Neural enhanced belief propagation for multiobject tracking},
  author={Liang, Mingchao and Meyer, Florian},
  journal={IEEE Transactions on Signal Processing},
  volume={72},
  pages={15--30},
  year={2023},
  publisher={IEEE}
}

@article{Suhail2021EnergyBasedLF,
  title={Energy-Based Learning for Scene Graph Generation},
  author={Mohammed Suhail and Abhay Mittal and Behjat Siddiquie and Chris Broaddus and Jayan Eledath and G{\'e}rard G. Medioni and Leonid Sigal},
  journal={2021 IEEE/CVF Conference on Computer Vision and Pattern Recognition (CVPR)},
  year={2021},
  pages={13931-13940},
  url={https://api.semanticscholar.org/CorpusID:232104772}
}

@InProceedings{Xu_2017_CVPR,
author = {Xu, Danfei and Zhu, Yuke and Choy, Christopher B. and Fei-Fei, Li},
title = {Scene Graph Generation by Iterative Message Passing},
booktitle = {Proceedings of the IEEE Conference on Computer Vision and Pattern Recognition (CVPR)},
month = {July},
year = {2017}
}

@InProceedings{Tang_2019_CVPR_VCTREE_mRaN,
author = {Tang, Kaihua and Zhang, Hanwang and Wu, Baoyuan and Luo, Wenhan and Liu, Wei},
title = {Learning to Compose Dynamic Tree Structures for Visual Contexts},
booktitle = {Proceedings of the IEEE/CVF Conference on Computer Vision and Pattern Recognition (CVPR)},
month = {June},
year = {2019}
}

@InProceedings{GraphRCNN,
author = {Yang, Jianwei and Lu, Jiasen and Lee, Stefan and Batra, Dhruv and Parikh, Devi},
title = {Graph R-CNN for Scene Graph Generation},
booktitle = {Proceedings of the European Conference on Computer Vision (ECCV)},
month = {September},
year = {2018}
}

@InProceedings{TransformerSGG1,
    author    = {Im, Jinbae and Nam, JeongYeon and Park, Nokyung and Lee, Hyungmin and Park, Seunghyun},
    title     = {EGTR: Extracting Graph from Transformer for Scene Graph Generation},
    booktitle = {Proceedings of the IEEE/CVF Conference on Computer Vision and Pattern Recognition (CVPR)},
    month     = {June},
    year      = {2024},
    pages     = {24229-24238}
}

@InProceedings{TransformerSGG2,
    author    = {Hayder, Zeeshan and He, Xuming},
    title     = {DSGG: Dense Relation Transformer for an End-to-end Scene Graph Generation},
    booktitle = {Proceedings of the IEEE/CVF Conference on Computer Vision and Pattern Recognition (CVPR)},
    month     = {June},
    year      = {2024},
    pages     = {28317-28326}
}

@InProceedings{DataResampling1,
    author    = {Li, Rongjie and Zhang, Songyang and Wan, Bo and He, Xuming},
    title     = {Bipartite Graph Network With Adaptive Message Passing for Unbiased Scene Graph Generation},
    booktitle = {Proceedings of the IEEE/CVF Conference on Computer Vision and Pattern Recognition (CVPR)},
    month     = {June},
    year      = {2021},
    pages     = {11109-11119}
}

@inproceedings{kim2024groupwise,
      title={Groupwise Query Specialization and Quality-Aware Multi-Assignment for Transformer-based Visual Relationship Detection}, 
      author={Kim, Jongha and Park, Jihwan and Park, Jinyoung and Kim, Jinyoung and Kim, Sehyung and Kim, Hyunwoo J},
      booktitle={CVPR},
      year={2024},
}

@inproceedings{WeiYang_ICLR2019,
    Author = {Wei Yang and Xiaolong Wang and Ali Farhadi and Abhinav Gupta and Roozbeh Mottaghi},
    Title = {Visual Semantic Navigation using Scene Priors},
    Booktitle = {ICLR},
    Year = {2019},
}

@article{Zhu2020HierarchicalPF,
  title={Hierarchical Planning for Long-Horizon Manipulation with Geometric and Symbolic Scene Graphs},
  author={Yifeng Zhu and Jonathan Tremblay and Stan Birchfield and Yuke Zhu},
  journal={2021 IEEE International Conference on Robotics and Automation (ICRA)},
  year={2020},
  pages={6541-6548},
  url={https://api.semanticscholar.org/CorpusID:229156165}
}

@inproceedings{conf/nips/HudsonM19,
  author = {Hudson, Drew A. and Manning, Christopher D.},
  booktitle = {NeurIPS},
  editor = {Wallach, Hanna M. and Larochelle, Hugo and Beygelzimer, Alina and d'Alché Buc, Florence and Fox, Emily B. and Garnett, Roman},
  ee = {http://papers.nips.cc/paper/8825-learning-by-abstraction-the-neural-state-machine},
  pages = {5901-5914},
  title = {Learning by Abstraction: The Neural State Machine.},
  url = {http://dblp.uni-trier.de/db/conf/nips/nips2019.html#HudsonM19},
  year = 2019
}

@article{Qian2023SceneGR,
  title={Scene Graph Refinement Network for Visual Question Answering},
  author={Tianwen Qian and Jingjing Chen and Shaoxiang Chen and Bo Wu and Yu-Gang Jiang},
  journal={IEEE Transactions on Multimedia},
  year={2023},
  volume={25},
  pages={3950-3961},
  url={https://api.semanticscholar.org/CorpusID:248350511}
}

@article{Kim2020HypergraphAN,
  title={Hypergraph Attention Networks for Multimodal Learning},
  author={Eun-Sol Kim and Woo-Young Kang and Kyoung-Woon On and Yu-Jung Heo and Byoung-Tak Zhang},
  journal={2020 IEEE/CVF Conference on Computer Vision and Pattern Recognition (CVPR)},
  year={2020},
  pages={14569-14578},
  url={https://api.semanticscholar.org/CorpusID:218881688}
}

@article{Wang2019CrossmodalSG,
  title={Cross-modal Scene Graph Matching for Relationship-aware Image-Text Retrieval},
  author={Sijin Wang and Ruiping Wang and Ziwei Yao and S. Shan and Xilin Chen},
  journal={2020 IEEE Winter Conference on Applications of Computer Vision (WACV)},
  year={2019},
  pages={1497-1506},
  url={https://api.semanticscholar.org/CorpusID:204402762}
}

@article{Johnson2018ImageGF,
  title={Image Generation from Scene Graphs},
  author={Justin Johnson and Agrim Gupta and Li Fei-Fei},
  journal={2018 IEEE/CVF Conference on Computer Vision and Pattern Recognition},
  year={2018},
  pages={1219-1228},
  url={https://api.semanticscholar.org/CorpusID:4593810}
}

@article{Dhamo2020SemanticIM,
  title={Semantic Image Manipulation Using Scene Graphs},
  author={Helisa Dhamo and Azade Farshad and Iro Laina and Nassir Navab and Gregory Hager and Federico Tombari and C. Rupprecht},
  journal={2020 IEEE/CVF Conference on Computer Vision and Pattern Recognition (CVPR)},
  year={2020},
  pages={5212-5221},
  url={https://api.semanticscholar.org/CorpusID:215415902}
}

@InProceedings{MDETR,
    author    = {Kamath, Aishwarya and Singh, Mannat and LeCun, Yann and Synnaeve, Gabriel and Misra, Ishan and Carion, Nicolas},
    title     = {MDETR - Modulated Detection for End-to-End Multi-Modal Understanding},
    booktitle = {Proceedings of the IEEE/CVF International Conference on Computer Vision (ICCV)},
    month     = {October},
    year      = {2021},
    pages     = {1780-1790}
}

@inproceedings{GLIP,
      title={Grounded Language-Image Pre-training},
      author={Liunian Harold Li* and Pengchuan Zhang* and Haotian Zhang* and Jianwei Yang and Chunyuan Li and Yiwu Zhong and Lijuan Wang and Lu Yuan and Lei Zhang and Jenq-Neng Hwang and Kai-Wei Chang and Jianfeng Gao},
      year={2022},
      booktitle={CVPR},
}

@article{GroundingDINO,
  title={Grounding dino: Marrying dino with grounded pre-training for open-set object detection},
  author={Liu, Shilong and Zeng, Zhaoyang and Ren, Tianhe and Li, Feng and Zhang, Hao and Yang, Jie and Li, Chunyuan and Yang, Jianwei and Su, Hang and Zhu, Jun and others},
  journal={arXiv preprint arXiv:2303.05499},
  year={2023}
}

@article{GroundingDINO1.5,
  title={Grounding dino 1.5: Advance the" edge" of open-set object detection},
  author={Ren, Tianhe and Jiang, Qing and Liu, Shilong and Zeng, Zhaoyang and Liu, Wenlong and Gao, Han and Huang, Hongjie and Ma, Zhengyu and Jiang, Xiaoke and Chen, Yihao and others},
  journal={arXiv preprint arXiv:2405.10300},
  year={2024}
}

@inproceedings{detclipv3,
  title={Detclipv3: Towards versatile generative open-vocabulary object detection},
  author={Yao, Lewei and Pi, Renjie and Han, Jianhua and Liang, Xiaodan and Xu, Hang and Zhang, Wei and Li, Zhenguo and Xu, Dan},
  booktitle={Proceedings of the IEEE/CVF conference on computer vision and pattern recognition},
  pages={27391--27401},
  year={2024}
}

@inproceedings{TRex2,
  title={T-rex2: Towards generic object detection via text-visual prompt synergy},
  author={Jiang, Qing and Li, Feng and Zeng, Zhaoyang and Ren, Tianhe and Liu, Shilong and Zhang, Lei},
  booktitle={European Conference on Computer Vision},
  pages={38--57},
  year={2024},
  organization={Springer}
}

@inproceedings{florence2,
  title={Florence-2: Advancing a unified representation for a variety of vision tasks},
  author={Xiao, Bin and Wu, Haiping and Xu, Weijian and Dai, Xiyang and Hu, Houdong and Lu, Yumao and Zeng, Michael and Liu, Ce and Yuan, Lu},
  booktitle={Proceedings of the IEEE/CVF Conference on Computer Vision and Pattern Recognition},
  pages={4818--4829},
  year={2024}
}

@article{GPT4o,
  title={Gpt-4o system card},
  author={Hurst, Aaron and Lerer, Adam and Goucher, Adam P and Perelman, Adam and Ramesh, Aditya and Clark, Aidan and Ostrow, AJ and Welihinda, Akila and Hayes, Alan and Radford, Alec and others},
  journal={arXiv preprint arXiv:2410.21276},
  year={2024}
}

@article{Hudson2019GQAAN,
  title={GQA: a new dataset for compositional question answering over real-world images},
  author={Drew A. Hudson and Christopher D. Manning},
  journal={ArXiv},
  year={2019},
  volume={abs/1902.09506},
  url={https://api.semanticscholar.org/CorpusID:67855531}
}

@article{COCOStuff,
  title={COCO-Stuff: Thing and Stuff Classes in Context},
  author={Holger Caesar and Jasper R. R. Uijlings and Vittorio Ferrari},
  journal={2018 IEEE/CVF Conference on Computer Vision and Pattern Recognition},
  year={2016},
  pages={1209-1218},
  url={https://api.semanticscholar.org/CorpusID:4396518}
}

@article{Objects365,
  title={Objects365: A Large-Scale, High-Quality Dataset for Object Detection},
  author={Shuai Shao and Zeming Li and Tianyuan Zhang and Chao Peng and Gang Yu and Xiangyu Zhang and Jing Li and Jian Sun},
  journal={2019 IEEE/CVF International Conference on Computer Vision (ICCV)},
  year={2019},
  pages={8429-8438},
  url={https://api.semanticscholar.org/CorpusID:207967883}
}

@article{Zhang2019BridgingTG,
  title={Bridging the Gap Between Anchor-Based and Anchor-Free Detection via Adaptive Training Sample Selection},
  author={Shifeng Zhang and Cheng Chi and Yongqiang Yao and Zhen Lei and Stan Z. Li},
  journal={2020 IEEE/CVF Conference on Computer Vision and Pattern Recognition (CVPR)},
  year={2019},
  pages={9756-9765},
  url={https://api.semanticscholar.org/CorpusID:208637257}
}

@article{Liu2021SwinTH,
  title={Swin Transformer: Hierarchical Vision Transformer using Shifted Windows},
  author={Ze Liu and Yutong Lin and Yue Cao and Han Hu and Yixuan Wei and Zheng Zhang and Stephen Lin and Baining Guo},
  journal={2021 IEEE/CVF International Conference on Computer Vision (ICCV)},
  year={2021},
  pages={9992-10002},
  url={https://api.semanticscholar.org/CorpusID:232352874}
}

@article{Lin2016FeaturePN,
  title={Feature Pyramid Networks for Object Detection},
  author={Tsung-Yi Lin and Piotr Doll{\'a}r and Ross B. Girshick and Kaiming He and Bharath Hariharan and Serge J. Belongie},
  journal={2017 IEEE Conference on Computer Vision and Pattern Recognition (CVPR)},
  year={2016},
  pages={936-944},
  url={https://api.semanticscholar.org/CorpusID:10716717}
}

@inproceedings{Vaswani2017AttentionIA,
  title={Attention is All you Need},
  author={Ashish Vaswani and Noam M. Shazeer and Niki Parmar and Jakob Uszkoreit and Llion Jones and Aidan N. Gomez and Lukasz Kaiser and Illia Polosukhin},
  booktitle={Neural Information Processing Systems},
  year={2017},
  url={https://api.semanticscholar.org/CorpusID:13756489}
}

@article{Kruskal1956,
 ISSN = {00029939, 10886826},
 URL = {http://www.jstor.org/stable/2033241},
 author = {Joseph B. Kruskal},
 journal = {Proceedings of the American Mathematical Society},
 number = {1},
 pages = {48--50},
 publisher = {American Mathematical Society},
 title = {On the Shortest Spanning Subtree of a Graph and the Traveling Salesman Problem},
 urldate = {2024-07-01},
 volume = {7},
 year = {1956}
}
}

\newpage
\renewcommand{\thesection}{\Alph{section}}
\setcounter{section}{0}

\section{Belief Propagation Algorithm}

\subsection{Algorithm Description}
\begin{figure}
  \centering
  \includegraphics[width=0.99\linewidth]{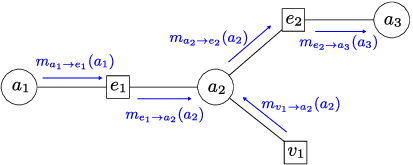}
  \caption{
    A simple MRF and the belief propagation algorithm.
    Circles represent variables, squares represent factors such as relationships or object labels, and blue arrows represent the flow of messages.
  }
  \label{fig:simple_mrf}
\end{figure}

Because belief propagation plays an important role in our method, we briefly explain its operation using the MRF in \Cref{fig:simple_mrf} as an example.
Suppose the probability distribution represented by the MRF is expressed by the following equation:
\begin{equation}
  \begin{aligned}
    &-\log P(a_1, a_2, a_3) \\
    &= v_1(a_2) + e_1(a_1, a_2) + e_2(a_2, a_3) + Z
    \label{eq:simple_mrf}
  \end{aligned}
\end{equation}

Belief propagation is a method of sending messages between the variables and factors of an MRF, as depicted in \Cref{fig:simple_mrf}, to obtain the marginal distributions and the normalization term $Z$.
The rules for sending these messages are as follows.
\begin{description}
  \item[Rule 1] A variable sends a specific factor the sum of all messages received from other factors.
  \item[Rule 2] A factor sends a specific variable the result of subtracting its own energy from the sum of the messages received from other variables and reducing the other variables by log-sum-exp.
  \item[Rule 3] A variable with only one connection sends 0, and a factor with only one connection sends the negative of its own energy.
  \item[Rule 4] The sum of all messages received from all connections, passed through an exponential function, is proportional to the marginal distribution of the variable.
\end{description}
Using these rules, we calculate the marginal distribution of $a_3$ as follows.
\begin{align}
  m_{a_1 \to e_1}(a_1) &= 0 \label{eq:bp1}\\
  m_{e_2 \to a_2}(a_2) &= \log \sum_{a_1} \exp(m_{a_1 \to e_1}(a_1) - e_1(a_1, a_2)) \label{eq:bp2} \\
  m_{v_1 \to a_2}(a_2) &= -v_1(a_2) \label{eq:bp3} \\
  m_{a_2 \to e_2}(a_2) &= m_{e_2 \to a_2}(a_2) + m_{v_1 \to a_2}(a_2) \label{eq:bp4} \\
  m_{e_2 \to a_3}(a_3) &= \log \sum_{a_2} \exp(m_{a_2 \to e_2}(a_2) - e_2(a_2, a_3)) \label{eq:bp5}
\end{align}
\Cref{eq:bp4} is the result of applying Rule 1, \Cref{eq:bp2,eq:bp5} are the results of applying Rule 2, and \Cref{eq:bp1,eq:bp3} are the results of applying Rule 3.
According to Rule 4, $\exp(m_{e_2 \to a_3}(a_3))$ is proportional to the marginal distribution of $a_3$.
This can be confirmed by substituting \Cref{eq:bp1,eq:bp2,eq:bp3,eq:bp4} and \Cref{eq:simple_mrf} into \Cref{eq:bp5}.
\begin{gather}
  \begin{aligned}
    &\exp(m_{e_2 \to a_3}(a_3)) \\
    &= \sum_{a_1, a_2} \exp(-v_1(a_2) - e_1(a_1, a_2) - e_2(a_2, a_3)) \\
    &= \sum_{a_1, a_2} \exp(\log P(a_1, a_2, a_3) + Z) \\
    &= \exp(Z) P(a_3)
  \end{aligned}
\end{gather}
This is proportional to the marginal distribution of $a_3$.
Considering that $\sum{a_3}P(a_3)=1$, we can calculate $Z$ as follows.
\begin{equation}
  Z = \log \sum_{a_3} \exp(m_{e_2 \to a_3}(a_3))
\end{equation}
Thus, we compute the marginal distribution of the variables and the normalization term $Z$ for training and inference. 
This algorithm involves only the log-sum-exp operation and summation, which are differentiable and numerically stable, thereby enabling stable backpropagation.

\subsection{Simplified Code}
This section presents a simplified module for the belief propagation algorithm in PyTorch and demonstrates its simplicity and auto-differentiability.
As a practical implementation note, we use \verb'weakref' to prevent potential memory leaks that can arise from circular references.

\begin{lstlisting}[language=Python]
# bp.py

from weakref import ref as wref
import torch


class Variable:
  def __init__(self, dim):
    self.dim = dim
    self.adjs = []

  def get_message(self, caller=None):
    x = 0.
    for f in self.adjs:
      if f is not caller:
        x += f.get_message(self)
    return x


class Factor:
  def __init__(self, energy, variables):
    self.energy = energy
    self.adjs = list(map(wref, variables))
    for v in variables:
      v.adjs.append(self)

  def get_message(self, caller):
    x = - self.energy
    for v in self.adjs:
      x.transpose_(0, -1)
      if v() is not caller:
        x += v().get_message(self)
        x = x.logsumexp(dim=-1)
    return x
\end{lstlisting}

To illustrate the usage of the module, we applied it to the MRF depicted in \cref{fig:simple_mrf}.
The following code demonstrates how to compute the marginal probability of variable $a_3$ the normalization term $Z$.

\begin{lstlisting}[language=Python]
# example.py

from bp import Variable, Factor
import torch

# Define variables
a1 = Variable(2)
a2 = Variable(3)
a3 = Variable(4)

# Define energies
e1 = torch.rand(2, 3)
e2 = torch.rand(3, 4)
v1 = torch.rand(3)

# Connect
Factor(e1, [a1, a2])
Factor(e2, [a2, a3])
Factor(v1, [a2])

a3_message = a3.get_message()

Z = a3_message.logsumexp(dim=0)
print(Z)

marginal_prob_a3 = torch.exp(a3_message - Z)
print(marginal_prob_a3)
\end{lstlisting}

As is evident from the code, the belief propagation algorithm is not only easy to implement, but also compatible with neural network frameworks.
We propose that belief propagation offers a viable approach for integrating graph structures into neural networks.

\section{Fourier Transform of Our Positional Encoding}
\begin{figure}[t]
  \centering
  \begin{subfigure}{0.35\linewidth}
    \includegraphics[bb=0 0 235 235, width=1.0\linewidth]{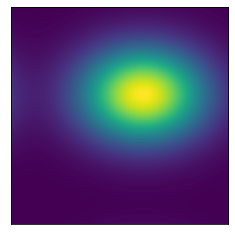}
    \captionsetup{format=hang, justification=raggedright}
    \caption{$(x, y, \sigma_x, \sigma_y) = (0.6, 0.4, 0.2, 0.15)$}
    \label{fig:posenc-large}
  \end{subfigure}
  \hspace{3em}
  \begin{subfigure}{0.35\linewidth}
    \includegraphics[bb=0 0 235 235, width=1.0\linewidth]{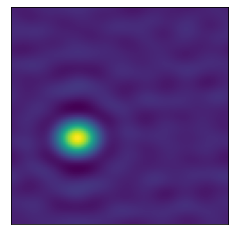}
    \captionsetup{format=hang, justification=raggedright}
    \caption{$(x, y, \sigma_x, \sigma_y) = (0.3, 0.6, 0.05, 0.03)$}
    \label{fig:posenc-small}
  \end{subfigure}
  \caption{
    Fourier transform results of our positional encoding.
    The y-axis increases from top to bottom.
    In the case of a small range (\Cref{fig:posenc-small}), a pattern appears in the background, which is the result of randomly omitting high-frequency components.
    }
  \label{fig:posenc}
\end{figure}

Our positional encoding method leverages Fourier transforms, enabling the encoding of positional relationships effectively. 
By applying an inverse Fourier transform to the encoding, we derive a function centered at $(x, y)$ with a width and height of $(\sigma_x, \sigma_y)$, as illustrated in \Cref{fig:posenc}, which captures the spatial range.
Additionally, the convolution property of Fourier transforms allows parallel movement and range expansion to be represented through simple multiplication operations. 
Furthermore, since the Fourier transform preserves inner products, the inner product of the encoded results reflects the inner product of the spatially spread functions, as shown in \Cref{fig:posenc}. 
It indicates the degree overlap between ranges.

\section{Object Grounding Results for VG-PO Dataset}

\begin{table}[h]
  \centering
  \small
  \begin{tabular}{l|rr|rr}
    \toprule
    Model & \multicolumn{2}{|c|}{Seen} & \multicolumn{2}{|c}{Unseen} \\
    & R@1 & R@5 & R@1 & R@5 \\
    \midrule
    Florence-2-L~\cite{florence2}          & 20.7 & - & 20.9 & - \\
    \midrule
    MDETR~\cite{MDETR}                     & 26.2 & 47.1 & 26.4 & 45.7 \\
    GLIP-T~\cite{GLIP}                     & 30.2 & 51.5 & 29.5 & 48.4 \\
    GroundingDINO~\cite{GroundingDINO}     & 32.1 & 36.9 & \textbf{31.6} & 34.6 \\
    \midrule
    VL-MPAG~\cite{VLMPAG} (w/o rel)        & 33.2 & 56.6 & 19.6 & 43.1 \\
    VL-MPAG~\cite{VLMPAG}                  & 39.9 & 66.9 & 29.0 & \textbf{53.6} \\
    SceneProp (Ours, w/o rel)              & 37.0 & 60.6 &  2.4 &  9.7 \\
    SceneProp (Ours)                       & \textbf{45.1} & \textbf{67.1} & 25.8 & 46.0 \\
    \bottomrule
  \end{tabular}
  \caption{
    The result of VG-PO dataset.
    "Seen" denotes the performance on the 125 object categories used for training, and "Unseen" denotes the performance on the 25 new object categories introduced during testing.
  }
  \label{tab:vgpo_results}
\end{table}

This section presents our results on the VG-PO (Visual Genome Partially Observed) dataset (\Cref{tab:vgpo_results}).
Derived from Visual Genome~\cite{Xu_2017_CVPR}, VG-PO is specifically designed to evaluate performance on unseen object categories and consists of 93K training and 40K test images.
Its key feature is the vocabulary split: 125 object categories are used for training, while the test set introduces an additional 25 new categories.

\begin{figure}
  \centering
  \includegraphics[width=0.99\linewidth]{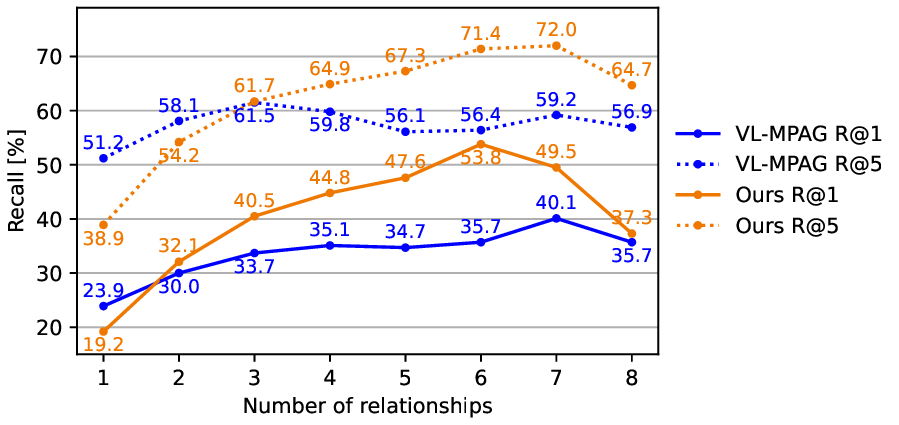}
  \caption{
    Impact of query graph size (number of relationships) on grounding performance on unseen category in VG-PO dataset. SceneProp's performance improves with more descriptive queries.
  }
  \label{fig:n_relation_vgpo}
\end{figure}

Although SceneProp is not explicitly designed for open-vocabulary detection, it can infer the location of unseen objects purely through their relational context.
This capability becomes more effective as queries become more descriptive.
As shown in \Cref{fig:n_relation_vgpo}, SceneProp's performance on unseen categories improves with the number of relationships in the query, eventually outperforming VL-MPAG on complex queries with more than two relationships.

Formally supporting open-vocabulary grounding, for instance by integrating a text encoder for category names, remains a promising direction for future work.

\section{Comparison with Scene Graph Generation}
In this section, we compare the tolerance to the long-tail distribution of relationship categories between scene graph grounding and scene graph generation.
We used SpeaQ~\cite{kim2024groupwise}, a state-of-the-art scene graph generation model, as a comparison target and evaluated its performance on the VG-150 dataset.
We used the trained model provided by SpeaQ.
We calculated the mean recall using the top k frequent relationship categories to show how the performance changed as the label distribution becames long-tailed.
Note that the scores cannot be directly compared because of the different tasks.
However, the comparison provides insights into the tolerance to the long-tail distributions.

\begin{figure}
  \centering
  \begin{subfigure}{0.99\linewidth}
    \includegraphics[width=0.80\linewidth]{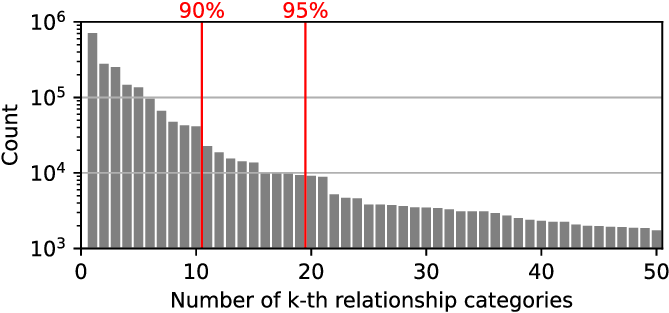}
    \caption{Relationship category distribution in VG-150 dataset.}
    \label{fig:relcat_count_vg150}
  \end{subfigure}
  \begin{subfigure}{0.99\linewidth}
    \includegraphics[width=0.99\linewidth]{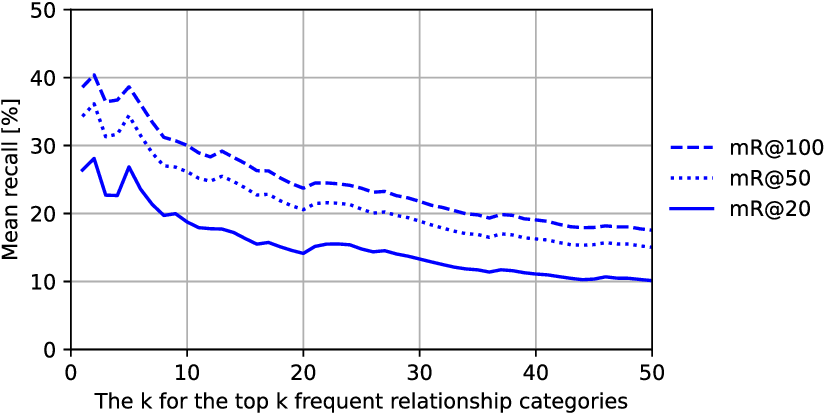}
    \caption{SpeaQ~\cite{kim2024groupwise} - Scene Graph Generation}
    \label{fig:speaq_recall_topk_freq_results_vg150}
  \end{subfigure}
  \begin{subfigure}{0.99\linewidth}
    \includegraphics[width=0.99\linewidth]{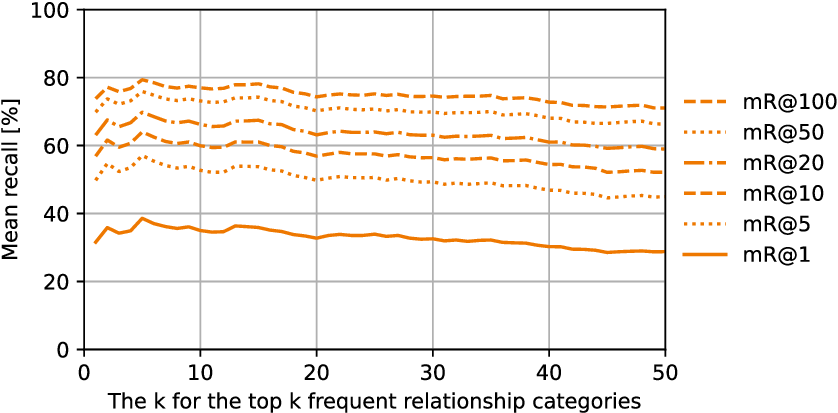}
    \caption{SceneProp - Scene Graph Grounding}
    \label{fig:sceneprop_recall_topk_freq_results_vg150}
  \end{subfigure}
  \caption{
    Mean recall for the top k frequent relationship categories on the VG-150 dataset.
    The x-axis shows the value of k.
    As k increases, the distribution becomes more long-tailed.
    The right end of the graph represents the mean recall for all relationship categories.
  }
  \label{fig:vg150_recall_topk_freq_results}
\end{figure}

\Cref{fig:relcat_count_vg150} shows the long-tailed distribution of relationship categories in the VG-150 dataset.
The top 10 and 19 categories accounted for 90\% and 95\% of the relationships, respectively.
\Cref{fig:speaq_recall_topk_freq_results_vg150} and \Cref{fig:sceneprop_recall_topk_freq_results_vg150} show the mean recall with the top k frequent relationship categories for SpeaQ and SceneProp, respectively.
The SpeaQ showed a significant performance drop as the label distribution became long-tailed, whereas SceneProp maintained stable performance.
Additionally, the recall (R@20, 50, and 100) for SpeaQ was 25.1, 32.1, and 35.5, respectively, whereas the mean recall (mR@20, 50, and 100) was 10.1, 15.0, and 17.5, respectively.
This significant difference between recall and mean recall indicates that SpeaQ performs poorly in the less-frequent relationship categories.
However, the differences in SceneProp between recall and mean recall are smaller than those in SpeaQ, as shown in \Cref{table:relation_pair_recall}, demonstrating stable performance even with a long-tailed category distribution.

We also evaluated the performance of SceneProp on the GQA dataset.
The result is shown in the main paper.

\section{Edge-removing Test for each number of relationships}
\begin{figure}
  \centering
  \includegraphics[width=0.99\linewidth]{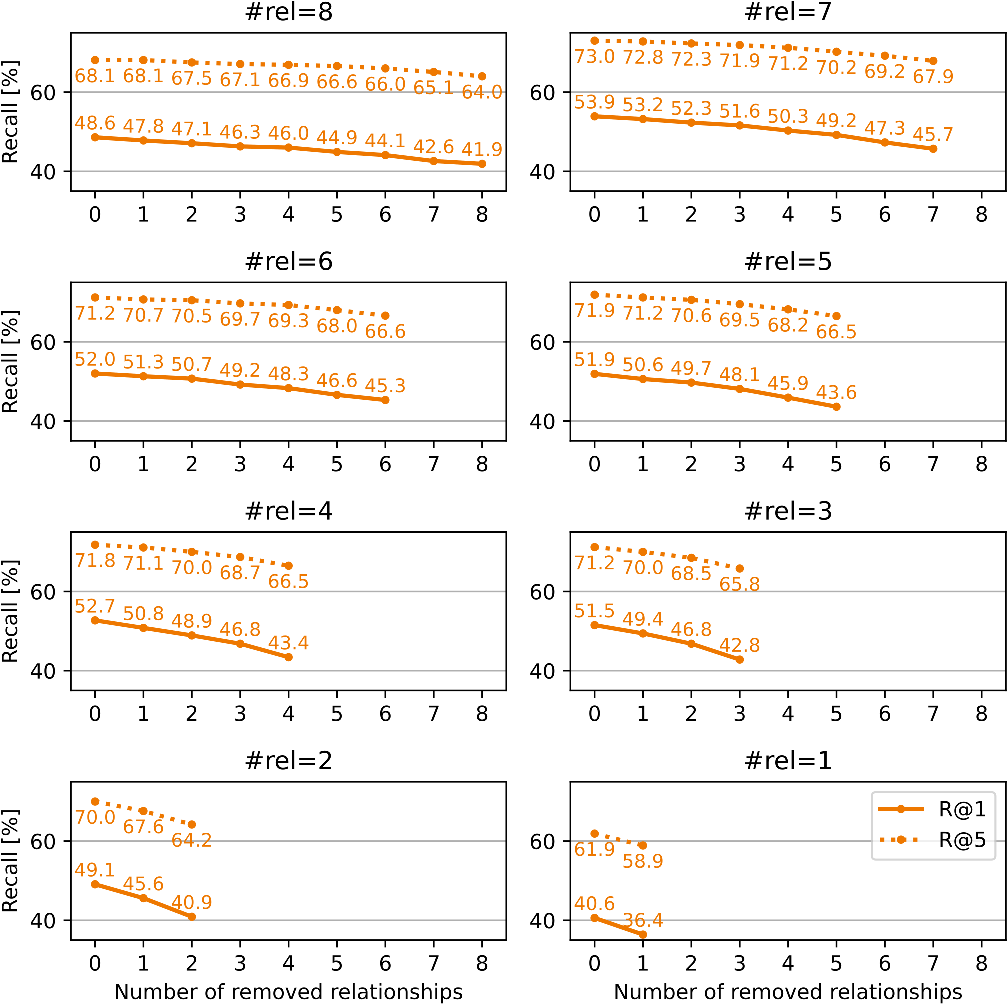}
  \caption{
    Performance on VG-FO dataset when randomly removing edges from the query graph.
    \#rels denotes the number of relationships (i.e., edges) in the original query graph.
  }
  \label{fig:removed_edge_test}
\end{figure}

In the main-text experiments, each query graph is paired with a different image and query difficulty varies across graphs, confounding the effect of the number of relationships.
To control for this, we evaluate SceneProp on variants of the \emph{same} query graph obtained by randomly deleting edges to reach specified \#rels while keeping nodes (and labels) fixed.
On VG-FO, \Cref{fig:removed_edge_test} shows a consistent drop in performance as edges are removed across all \#rels buckets, supporting the claim that additional relations provide informative constraints that SceneProp effectively exploits.

\section{Generalization to Cyclic Query Graphs}

\begin{table}
  \centering
  \begin{tabular}{l|rr|rr}
    \toprule
    \multirow{2}{*}{\#rels}  & \multicolumn{2}{|c|}{w/o loop} & \multicolumn{2}{|c}{w/ loop} \\
     & R@1 & R@5 & R@1 & R@5 \\
    \midrule
    1 & 40.6 & 61.9 & - & - \\
    2 & 48.9 & 69.6 & 53.9 & 76.1 \\
    3 & 50.8 & 70.7 & 59.3 & 79.2 \\
    4 & 51.4 & 70.8 & 60.9 & 79.1 \\
    5 & 50.4 & 70.7 & 57.8 & 77.8 \\
    6 & 50.0 & 69.3 & 60.0 & 78.6 \\
    7 & 51.2 & 70.9 & 60.8 & 78.2 \\
    8 & 46.7 & 67.0 & 53.4 & 72.0 \\
    \midrule
    all & 45.7 & 66.3 & 58.1 & 77.6 \\
    \bottomrule
  \end{tabular}
  \caption{
    Performance on acyclic (w/o loop) versus cyclic (w/ loop) query graphs in the VG-FO dataset.
    \#rels denotes the number of relationships (i.e., edges) in the query graph.
  }
  \label{table:loop_results}
\end{table}

We report the full numeric breakdown on VG-FO by the number of relationships (\#rels) in the query graph; see \Cref{table:loop_results} for acyclic (w/o loop) versus cyclic (w/ loop) results.
Algorithmic details on handling cycles (loopy belief propagation with a brief MCMC refinement) are given in the main text.
Across most \#rels and in the overall average, cyclic graphs outperform acyclic ones, consistent with the view that loops provide richer relational constraints (i.e., more relations per object) that SceneProp can leverage; note that loops are undefined at \#rels=1, hence the “–” entries.

\section{Additional Relationship Pair Grounding Results}
This section presents the relationship pair grounding results for the VG-FO, VG-PO, COCOStuff, and GQA datasets, as shown in \Cref{table:relation_pair_recall}.
Because of the difficulty in predicting relation pairs, we also provided the top 10, 20, 50, and 100 recalls, in addition to the top 1 and 5 recalls.

\begin{table}[h]
  \centering
  \begin{tabular}{l|rrrrr}
    \toprule
     & VG- & VG- & COCO- & \multirow{2}{*}{GQA} & VG- \\
     &  FO &  PO & Stuff &                      & 150 \\
    \midrule
    R@1    & 31.3 & 14.6 & 46.8 & 32.5 & 35.9 \\
    R@5    & 46.6 & 27.7 & 70.0 & 46.7 & 54.1 \\
    R@10   & 53.4 & 35.1 & 77.3 & 52.2 & 61.2 \\
    R@20   & 59.7 & 42.8 & 82.8 & 56.8 & 67.2 \\
    R@50   & 66.9 & 53.1 & 88.2 & 61.7 & 73.7 \\
    R@100  & 71.3 & 60.0 & 91.5 & 64.4 & 77.4 \\
    \midrule
    mR@1   & 27.9 & 14.9 & 48.5 & 39.0 & 28.8 \\
    mR@5   & 43.1 & 26.6 & 71.0 & 53.2 & 44.9 \\
    mR@10  & 49.9 & 32.6 & 77.6 & 59.2 & 52.2 \\
    mR@20  & 56.5 & 40.2 & 83.4 & 65.0 & 59.0 \\
    mR@50  & 63.8 & 50.7 & 88.8 & 69.8 & 66.3 \\
    mR@100 & 68.6 & 56.9 & 92.1 & 72.8 & 71.1 \\
    \bottomrule
  \end{tabular}
  \caption{Relationship-pair grounding performance on VG-FO, VG-PO, COCOStuff, VG-150, and GQA datasets.}
  \label{table:relation_pair_recall}
\end{table}

\section{Sensitivity analysis of the hyperparameters}
This section presents a the sensitivity analysis of the hyperparameters in SceneProp.
We evaluated the performance of SceneProp with different hyperparameters using the VG-FO dataset.

First, we analyzed the sensitivity of the number of layers in the Transformer within the relationship feature extractor.
\Cref{fig:sensitivity_num_layers} shows the recall at top-1 and top-5 with different numbers of layers.
Performance was not significantly affected by the number of layers; however, two layers showed the best performance.

\begin{table}[h]
  \centering
  \begin{tabular}{l|rrrrr}
    \toprule
    n layers & 0 & 1 & 2 & 3 & 4 \\
    \midrule
    R@1    & 44.2 & 44.7 & \textbf{46.6} & 44.6 & 43.0 \\
    R@5    & 65.1 & 65.3 & \textbf{67.1} & 65.6 & 64.6 \\
    \bottomrule
  \end{tabular}
  \caption{Sensitivity analysis of the number of layers in the Transformer using the VG-FO dataset.}
  \label{fig:sensitivity_num_layers}
\end{table}

Next, we analyzed the sensitivity of the number of candidate bounding boxes.
\Cref{fig:sensitivity_num_candidates} shows the recall at top-1 and top-5 with different numbers of candidate bounding boxes.
The number of candidate bounding boxes significantly affected the performance, with 512 candidates showing the best results.
The decrease in performance with fewer candidates was owing to the lack of candidate bounding boxes for correct grounding.

\begin{table}[h]
  \centering
  \begin{tabular}{l|rrrr}
    \toprule
    n candidates & 256 & 512 & 786 & 1024 \\
    \midrule
    R@1    & 29.0 & \textbf{46.6} & 45.6 & 44.3 \\
    R@5    & 47.9 & \textbf{67.1} & 66.6 & 66.2 \\
    \bottomrule
  \end{tabular}
  \caption{Sensitivity analysis of the number of candidate bounding boxes using the VG-FO dataset.}
  \label{fig:sensitivity_num_candidates}
\end{table}

\section{Unary vs.\ pairwise contributions.}

\begin{table}
  \centering
  \small
  \begin{tabular}{l|rr|rr|rr}
    \toprule
     & \multicolumn{2}{|c|}{full} & \multicolumn{2}{|c|}{w/o rels} & \multicolumn{2}{|c}{w/o objs} \\
     & & & \multicolumn{2}{|c|}{(unary only)} & \multicolumn{2}{|c}{(pairwise only)} \\
     & R@1 & R@5 & R@1 & R@5 & R@1 & R@5 \\
    \midrule
    VG-FO      & 46.6 & 67.1 & 40.0 & 64.6 & 26.0 & 46.6 \\
    VG150      & 43.7 & 68.4 & 37.4 & 64.1 & 24.9 & 43.5 \\
    COCO-Stuff & 68.2 & 88.9 & 48.1 & 82.8 & 51.3 & 81.1 \\
    GQA        & 53.6 & 68.2 & 33.4 & 54.5 & 21.3 & 36.1 \\
    \bottomrule
  \end{tabular}
  \caption{
    Performance comparison of that using both unary and pairwise terms (full), only unary terms (w/o rels), and only pairwise terms (w/o objs).
  }
  \label{table:unary_vs_pairwise}
\end{table}

In this section, we analyze the contributions of unary and pairwise energies in our model.
\Cref{table:unary_vs_pairwise} reports performance when using both unary and pairwise terms (full), only unary terms (w/o rels), and only pairwise terms (w/o objs).
On VG-FO, VG-150, and GQA, performance is significantly higher with only unary terms than with only pairwise terms.
This indicates that unary energies carry the larger share of evidence, consistent with the intuition that relations alone are insufficient for reliable grounding.
However, on COCO-Stuff, performance with only pairwise terms is comparable to that with only unary terms.
This likely reflects that COCO-Stuff is synthesized from object coordinates, making relative positions more informative than in the other datasets.

\section{Additional Visual Examples}
\begin{figure*}
  \centering
  \includegraphics[width=0.99\linewidth]{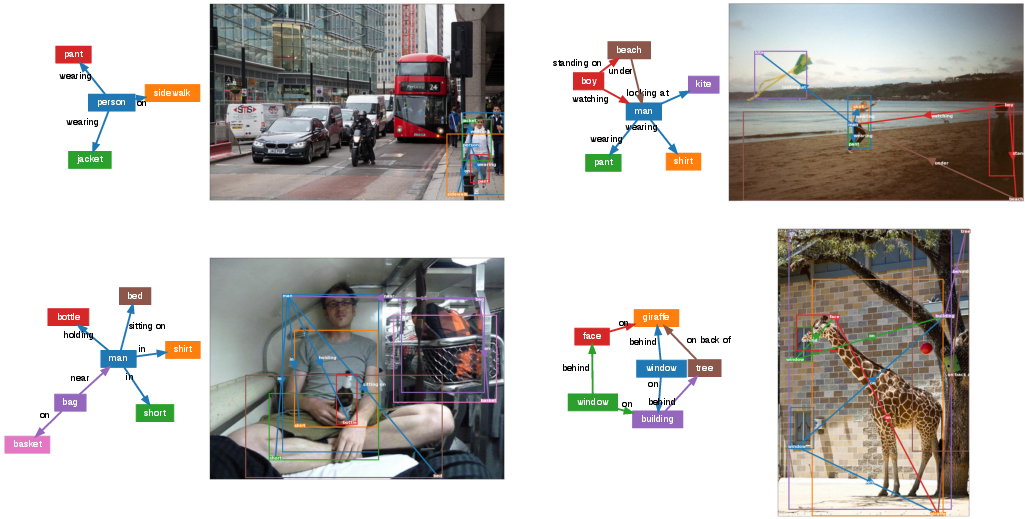}
  \caption{
    Grounding results on the VG-FO dataset.
    The right side of the query graph displays the grounding result.
    The colors of the nodes in the query graph match the colors of the bounding boxes in the grounding image.
  }
  \label{fig:vgfo_examples}
\end{figure*}

\begin{figure*}
  \centering
  \includegraphics[width=0.99\linewidth]{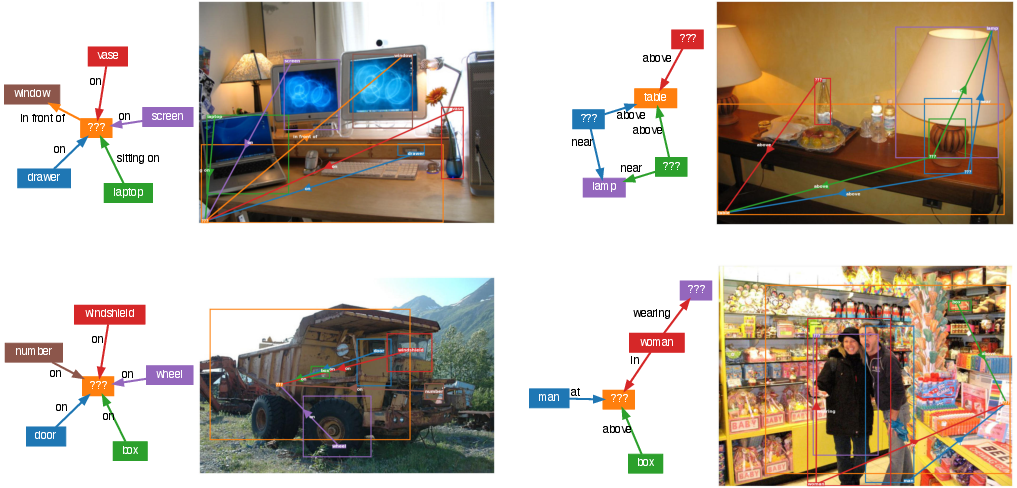}
  \caption{
    Grounding results on the VG-PO dataset.
    "???" denotes object categories absent from the training set.
  }
  \label{fig:vgpo_examples}
\end{figure*}

\begin{figure*}
  \centering
  \includegraphics[width=0.99\linewidth]{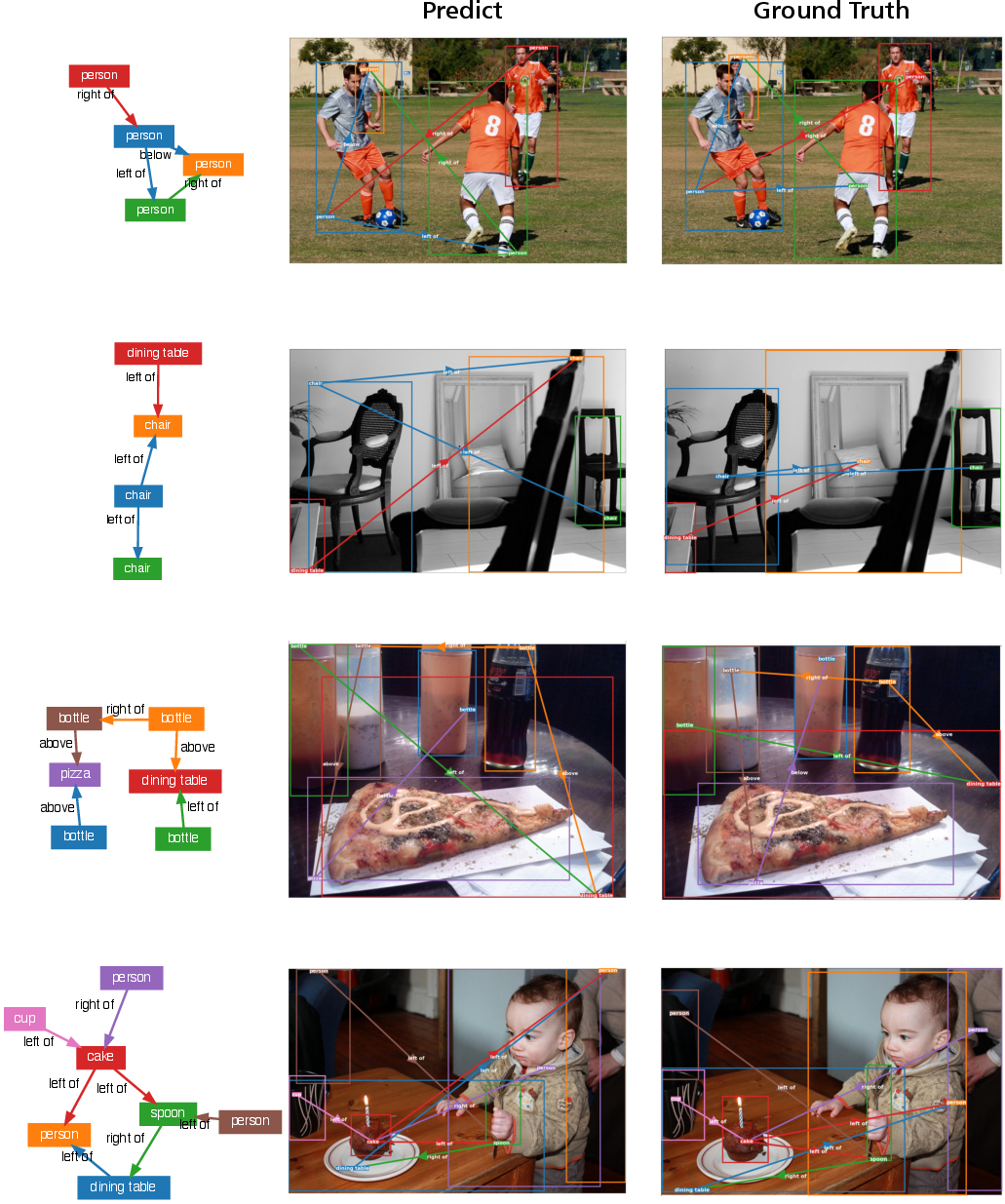}
  \caption{
    Grounding results on the COCOStuff dataset.
    Given the non-intuitive nature of scene graphs in the COCOStuff dataset, we include the ground truth images.
    The images on the right are the ground truth.
  }
  \label{fig:coco_examples}
\end{figure*}

\begin{figure*}
  \centering
  \includegraphics[width=0.90\linewidth]{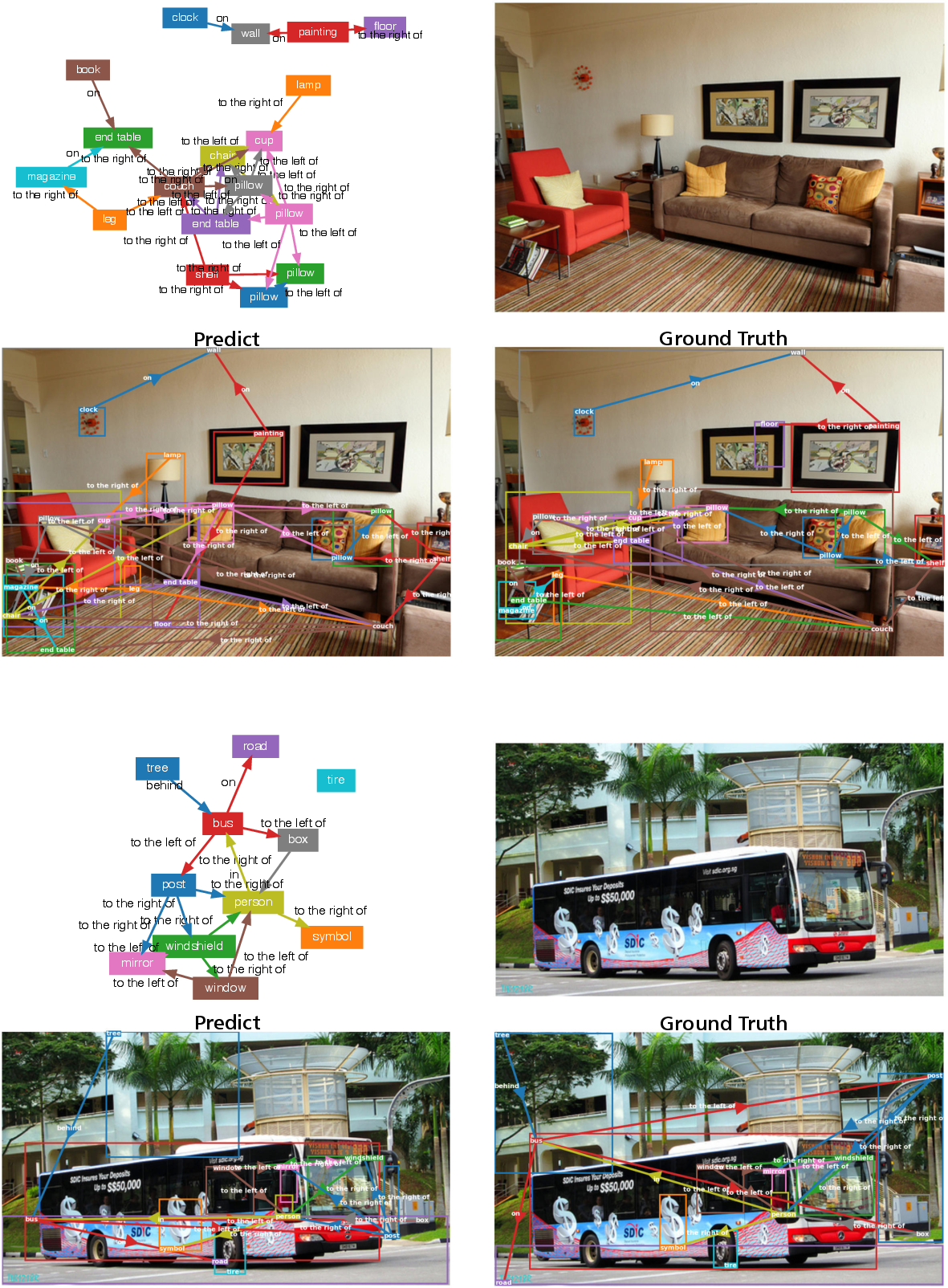}
  \caption{
    Grounding results on the GQA dataset.
    To facilitate better visibility of the original images, obscured by numerous bounding boxes, we also present images without bounding boxes.
    The lower left image represents the prediction result, while the lower right image is the ground truth.
  }
  \label{fig:gqa_examples}
\end{figure*}

\begin{figure*}
  \centering
  \includegraphics[width=0.90\linewidth]{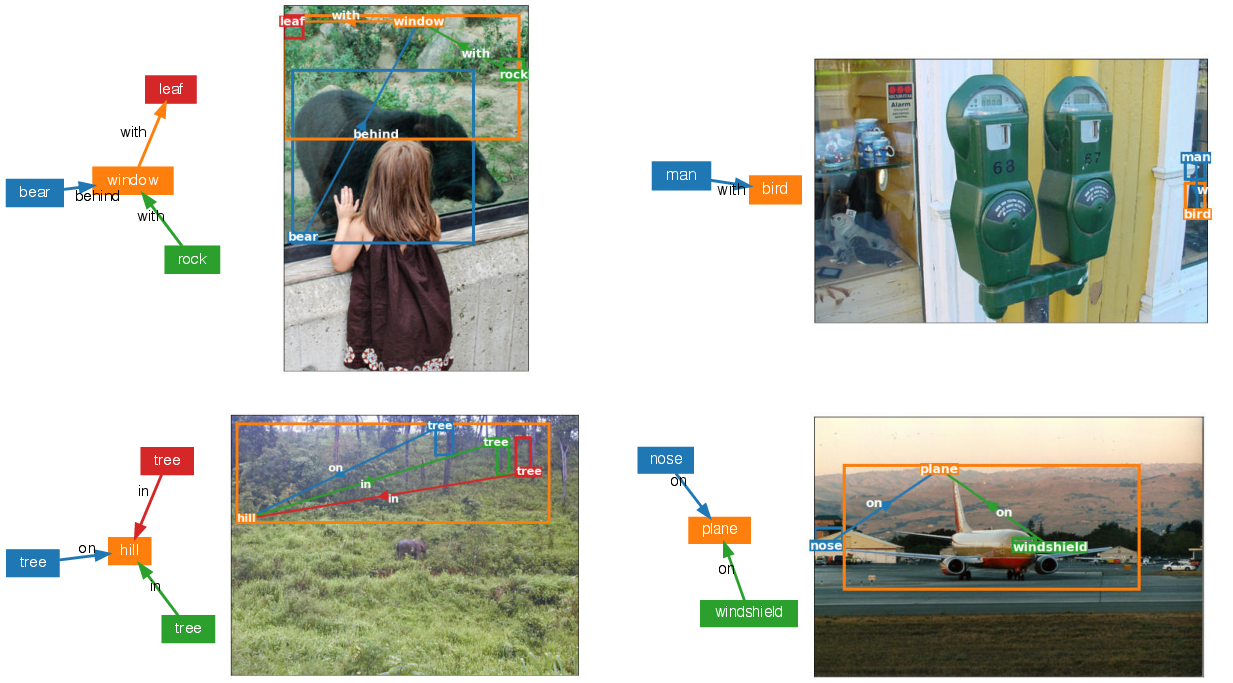}
  \caption{
    Examples of failure cases in SceneProp.
    Failures typically involve either (1) inaccurate localization for objects with ambiguous boundaries (left), or (2) incorrect grounding due to challenging visual conditions (right).
    In the top right example, the query asked for a "man" and a "bird" on the upper level of the left-side showcase, but these objects are small and have significant reflections, leading to incorrect grounding.
    In the bottom right example, the grounding failure appears to stem from the challenge of semantically associating the concept of a "nose" with a "plane."
  }
  \label{fig:sceneprop_failure_examples}
\end{figure*}

Examples of the grounding results for the VG-FO, VG-PO, COCOStuff, and GQA datasets are provided in \cref{fig:vgfo_examples}, \cref{fig:vgpo_examples}, \cref{fig:coco_examples}, and \cref{fig:gqa_examples}.
\Cref{fig:sceneprop_failure_examples} shows examples of failure cases in SceneProp.

\newpage

\end{document}